\documentclass[letterpaper]{article} 
\usepackage[submission]{aaai2026}  
\usepackage{times}  
\usepackage{helvet}  
\usepackage{courier}  
\usepackage[hyphens]{url}  
\usepackage{graphicx} 
\usepackage{natbib}  
\usepackage{caption} 
\usepackage{algorithm}
\usepackage{algorithmic}
\usepackage{amsmath}
\usepackage{amsfonts}
\usepackage{amssymb}
\usepackage{bbold}
\usepackage{multirow}
\usepackage{subfigure}

\usepackage{newfloat}
\usepackage{listings}
\DeclareCaptionStyle{ruled}{labelfont=normalfont,labelsep=colon,strut=off} 
\floatstyle{ruled}
\newfloat{listing}{tb}{lst}{}
\floatname{listing}{Listing}
\title{X2Edit: Revisiting Arbitrary-Instruction Image Editing through Self-Constructed Data and Task-Aware Representation Learning}

\author{
    Jian Ma\textsuperscript{\rm 1}\thanks{Co-first authors.},
    Xujie Zhu\textsuperscript{\rm 2}\footnotemark[1]\thanks{The author did his work during internship at OPPO AI Center.},
    Zihao Pan\textsuperscript{\rm 2}\footnotemark[2],
    Qirong Peng\textsuperscript{\rm 1},
    Xu Guo\textsuperscript{\rm 3}\footnotemark[2],
    Chen Chen\textsuperscript{\rm 1},
    Haonan Lu\textsuperscript{\rm 1}
}
\affiliations{
    \textsuperscript{\rm 1}OPPO AI Center,
    \textsuperscript{\rm 2}Sun Yat-sen University,
    \textsuperscript{\rm 3}Tsinghua University\\
    majian2@oppo.com, zhuxj6@mail2.sysu.edu.cn, panzh33@mail2.sysu.edu.cn, pengjirong@oppo.com, guo-x24@mails.tsinghua.edu.cn, chenchen4@oppo.com, luhaonan@oppo.com
    
}

\usepackage{bibentry}

\begin{document}

\maketitle
\begin{abstract}
Existing open-source datasets for arbitrary-instruction image editing remain suboptimal, while a plug-and-play editing module compatible with community-prevalent generative models is notably absent. In this paper, we first introduce the X2Edit Dataset, a comprehensive dataset covering 14 diverse editing tasks, including subject-driven generation. We utilize the industry-leading unified image generation models and expert models to construct the data. Meanwhile, we design reasonable editing instructions with the VLM and implement various scoring mechanisms to filter the data. As a result, we construct 3.7 million high-quality data with balanced categories. Second, to better integrate seamlessly with community image generation models, we design task-aware MoE-LoRA training based on FLUX.1, with only 8\% of the parameters of the full model. To further improve the final performance, we utilize the internal representations of the diffusion model and define positive/negative samples based on image editing types to introduce contrastive learning. Extensive experiments demonstrate that the model's editing performance is competitive among many excellent models. Additionally, the constructed dataset exhibits substantial advantages over existing open-source datasets. The open-source code, checkpoints, and datasets for X2Edit can be found at the following link: https://github.com/OPPO-Mente-Lab/X2Edit.
\end{abstract}

\begin{figure*}[htbp]
\centering
\includegraphics[width=1\textwidth]{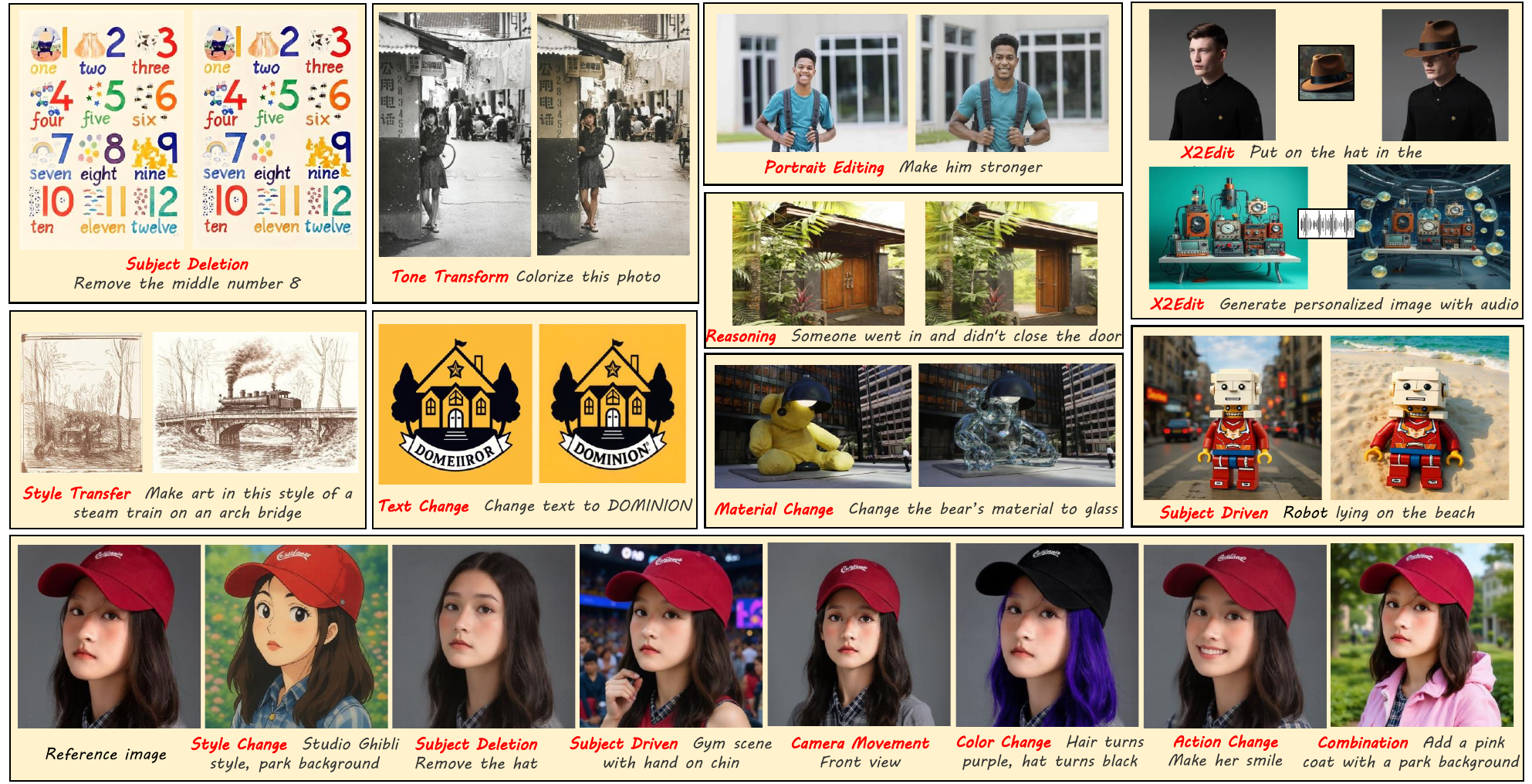}
\caption{The X2Edit image generation results span 14 diverse editing types. In each unit, the left image serves as the reference. The central modality in the top-right unit is the input to X2I and can be leveraged by other modalities to assist in image editing.
}
\label{Fig:1}
\end{figure*}

\section{Introduction}

Instruction-based image editing has witnessed explosive growth recently, evolving from single-task frameworks to more flexible models capable of open-vocabulary and free-form editing. Open-source models\cite{wang2025seededit,openai2024gpt4ocard,geminiteam2025geminifamilyhighlycapable} continue to lag behind their closed-source counterparts. Despite a significant increase in the release of editing datasets, the challenge of constructing a well-balanced, high-quality dataset that encompasses a wide range of editing tasks remains unresolved.

Existing editing datasets have three main limitations: \textbf{cumbersome construction processes}, \textbf{poor data quality}, and \textbf{limited support for complex editing tasks}. (1) Editing datasets such as AnyEdit\cite{yu2025anyeditmasteringunifiedhighquality} and ImgEdit\cite{ye2025imgeditunifiedimageediting} require different data construction processes to be designed for each type of editing task, which not only consume substantial human effort but is also difficult to scale flexibly. (2) Existing open-source datasets perform poorly in terms of editing accuracy and data balance. Datasets like AnyEdit and SEED-Data-Edit\cite{ge2024seeddataedittechnicalreporthybrid} suffer from low image quality and imbalanced data across different editing tasks. HQ-Edit\cite{hui2024hq} and OmniEdit\cite{wei2025omnieditbuildingimageediting} extensively use synthetic images as source images in hopes of improving quality, but this creates deviations from real data distributions. Moreover, existing workflows that leverage tools such as Flux-Fill\cite{flux2024} and SD 3.5\cite{esser2024scalingrectifiedflowtransformers} ControlNet necessitate instruction conversion and the acquisition of additional prompt information, both of which can readily result in deviations from the intended edits. (3) High-quality open-source data for difficult editing tasks involving complex reasoning, camera movement, style transfer, etc., is extremely scarce due to the difficulty of construction. 

In this paper, we propose an automated dataset construction pipeline and a large-scale training dataset called \textbf{X2Edit Dataset}. We uniformly use Vision-Language Model(VLM) to generate quantity-balanced instructions for different editing tasks based on source images and contextual examples. Subsequently, we leverage SOTA open-source and closed-source models to create editing pairs according to task-specific workflows. Finally, we introduce a comprehensive filtering mechanism, including pre-filtering of source images and further screening after comprehensive scoring, strictly ensuring data quality. Through this pipeline, we ultimately obtain a category-balanced, high-quality dataset of \textbf{3.7M} scale, which demonstrates strong competitiveness compared to existing open-source datasets in terms of data scale, number of supported tasks, and data quality.

Parallel to this, the rapid emergence of arbitrary-instruction image editing methods has dramatically advanced text-guided visual manipulation. However, these gains are often offset by prohibitive training costs. Step1X-Edit\cite{liu2025step1xeditpracticalframeworkgeneral} and Kontext\cite{labs2025flux1kontextflowmatching}, fine-tune the entire 12B-parameter DiT\cite{peebles2023scalable} backbone, whereas unified methods such as Bagel\cite{deng2025emergingpropertiesunifiedmultimodal} and OmniGen2\cite{wu2025omnigen2explorationadvancedmultimodal} develop cross-modal understanding and generation capabilities from scratch on massive multimodal corpora. HiDream-E\cite{hidreami1technicalreport} follows the same recipe of end-to-end pre-training with a 17B-parameter model on ultra-large-scale data. ICEdit\cite{zhang2025incontexteditenablinginstructional} inserts Mixture-of-Experts(MoE)\cite{shazeer2017outrageouslylargeneuralnetworks} layers into LoRA modules based on FLUX-Fill. Despite effectively reducing training costs, it is limited by transferability and still falls short of full-model methods in terms of editing fidelity.

In this paper, we present \textbf{X2Edit}. Specifically, we first learn a task embedding matrix whose entries are injected into the MoE gating network to guide expert selection. We also analogize different editing tasks within the same batch to negative samples and the same editing task to positive samples in contrastive learning. This division can promote the mapping of different editing tasks to distinct, separable locations in the hidden space projection, enabling the model to learn discriminative features and avoid feature collapse. In addition, it can also ensure that the same editing task is mapped to similar encodings in the hidden space projection. Subsequently, through the constraint of the contrastive regularization loss, X2Edit achieves further enhancement in overall performance. Our contributions are three-fold:
\begin{itemize}
    \item We construct X2Edit Dataset and GEdit-Bench++.
    \item To the best of our knowledge, X2Edit is an early attempt to explore the use of contrastive learning in arbitrary-instruction image editing.
    \item Extensive evaluations demonstrate that X2Edit Dataset surpasses existing open-source datasets across multiple objective metrics, while X2Edit rivals current SOTA editing methods in both automatic and human assessments.
\end{itemize}

\section{Related Work}

\subsection{Datasets for Image Editing}

\begin{table}[ht!]
\centering
\footnotesize
\begingroup
\setlength{\tabcolsep}{2.5pt}
\renewcommand{\arraystretch}{0.8}
\begin{tabular}{l|ccc|c}
\hline
\multirow{2}{*}{Datasets} & \multirow{2}{*}{\#Size} & \multirow{2}{*}{\#Types} & \multirow{2}{*}{Res.(px)} & \multirow{2}{*}{Complex Tasks} \\
 &  &  &  &  \\ \hline
AnyEdit & 2.5M & 25 & 512 & $\checkmark$ \\
HQ-Edit & 197K & 6 & $\geq$ 768 & $\times$ \\
UltraEdit & 4M & 9 & 512 & $\times$ \\
SEED-Data-Edit & 3.7M & 6 & 768 & $\times$ \\
ImgEdit & 1.2M & 13 & $\geq$ 1280 & $\times$ \\ 
OmniEdit & 5.2M/1.2M & 7 & $\geq$ 512 & $\times$ \\ \hline
\textbf{X2Edit(512)} & 2M & 14 & 512 & $\checkmark$ \\ 
\textbf{X2Edit(1024)} & 1.7M & 14 & $\sim$1024 & $\checkmark$ \\
\hline
\end{tabular}
\caption{Comparison of existing image editing datasets.}
\label{Tab: dataset1}
\endgroup
\end{table}

We compare with current representative open-source arbitrary-instruction  image editing datasets in Table~\ref{Tab: dataset1}. Represented by HQ-Edit, AnyEdit and OmniEdit, most existing datasets use automated pipelines to scale up as much as possible, while SEED-Data-Edit and UltraEdit\cite{zhao2024ultraeditinstructionbasedfinegrainedimage} add some manual quality control. The source images of HQ-Edit mostly come from synthetic data, causing the dataset to deviate from real-world images. Existing datasets also face problems of few editing categories and data imbalance. Although AnyEdit and ImgEdit enrich editing tasks, the overall pipeline suffers from critical flaws including excessive complexity, poor replicability and low data quality. Moreover, existing image editing datasets often do not include some complex editing tasks such as reasoning, subject-driven generation and style transfer. While ensuring high-quality data construction, we streamline the pipeline as much as possible, using GPT-4o, BAGEL, and Kontext to supplement data for complex editing tasks.

\subsection{Models for Image Editing}

Arbitrary-instruction image editing is advancing along three primary paradigms. The first is full-parameter training, which encompasses unified vision editing models and unified understanding-generation models. In the former camp, Kontext and Hidream-E train large-scale networks on vast datasets, whereas Step1X-Edit performs full parameter fine-tuning on the entire DiT framework. In the latter camp, Bagel and OmniGen2 pursue a unified understanding-generation model via joint pre-training on massive data, yet incur prohibitive compute and training costs. For AnyEdit\cite{yu2025anyeditmasteringunifiedhighquality}, the first training stage involves pre-training the UNet backbone of the diffusion process, while the second stage separately trains IP-Adapter as task-specific expert with MoE layers. The second is parameter-efficient fine-tuning to drastically cut training costs. ICEdit integrates MoE layers within LoRA modules for a lightweight approach. UniControl\cite{qin2023unicontrolunifieddiffusionmodel} introduces a task-aware HyperNet to modulate the diffusion models. The last is the optimization strategies for inference. These strategies\cite{hertz2022prompt,cao2023masactrl,wei2025freeflux,hu2025dcedit,feng2025personalize,li2025editid,zhu2025kvedittrainingfreeimageediting} completely avoid model training and instead guide the editing process by manipulating the intermediate states of the denoising process during inference. Conversely, while contrastive learning has proven effective for discriminative tasks, its application in generative models remains a nascent yet promising direction. REPA\cite{yu2025representationalignmentgenerationtraining} leverages external alignment for semantic inheritance at higher costs, whereas Dispersive Loss\cite{wang2025diffusedisperseimagegeneration} optimizes internal representations without external dependencies, both enhancing text-to-image generation.

\section{X2Edit Dataset}
We propose \textbf{X2Edit Dataset}, a large-scale, high-quality, and well-balanced dataset tailored for arbitrary-instruction image editing, with its construction process depicted in Figure~\ref{Fig: dataset pipeline}. Additional details can be found in the appendix D.

\begin{figure*}[htbp]
\centering
\includegraphics[width=1\textwidth]{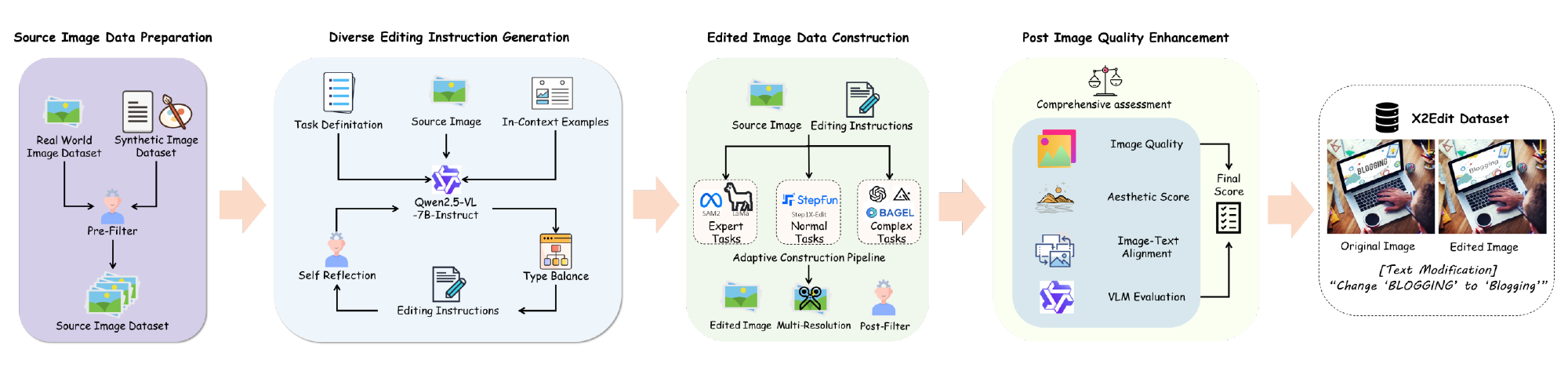}
\caption{The comprehensive construction pipeline of X2Edit Dataset. We divide the pipeline into four stages: (1) Sampling from real-world datasets and synthesizing source images using our internal query dataset; (2) Generating diverse editing instructions using a VLM based on the source images; (3) Generating edited images using task-specific workflows according to the editing instructions; (4) Conducting comprehensive evaluation and filtering of all generated data to ensure quality.
}
\label{Fig: dataset pipeline}
\end{figure*}

\subsection{Edit Type Definition}
To adapt to the diverse image editing needs of various users and settings, we categorize 14 distinct editing tasks. Figure~\ref{Fig: dataset statistic} provides details on these task categories and their statistical data. The comprehensive set includes local, global, and complex editing tasks, as well as subject-driven generation. Local editing focuses on modifications limited to particular regions of an image, leaving other areas unchanged, such as adding subjects or altering colors. Global editing refers to changes that affect the entire content of the image. Notably, complex editing tasks like reasoning editing and camera movement require models to thoroughly comprehend both the instructions and the source images. Furthermore, we introduce subject-driven generation, incorporating style transfer based on reference image, which are seldom featured in current open-source datasets.

\subsection{Automatic Dataset Pipeline}

\textbf{Source Image Preparation}. We select the majority of the source images from COYO-700M\cite{kakaobrain2022coyo-700m}, Wukong\cite{gu2022wukong} and LAION\cite{schuhmann2022laion5bopenlargescaledataset}, aiming to thoroughly encompass a diverse range of real-world image inputs. In order to fulfill the need for high-quality reference image in subject-driven generation, we also employ an internal query to generate source images using Shuttle-3-Diffusion\cite{shuttle3diffusion}. Rigorous filtering criterias are applied to samples that advance to the subsequent phase of the data construction pipeline. Specifically, we only retain images with high aesthetic score and minimum side lengths greater than 512 pixels. Furthermore, we filter the internal query for subject-driven generation with Qwen3\cite{yang2025qwen3technicalreport} to ensure the presence of keywords related to the subject in the foreground. Additionally, the data we construct encompasses a range of aspect ratios between 512 and 2048 to enhance diversity.

\textbf{Diverse Editing Instruction Generation}. Since image captions cannot capture rich visual detail information, we employ Qwen2.5-VL-7B\cite{bai2025qwen2} to directly generate diverse editing instructions based on source images, which differs from existing LLM-based methods. We provide source images, task definitions, and contextual examples, and use meticulously crafted prompts to guide the VLM in formulating editing instructions directly from the images. To minimize the risk that the VLM generates unfeasible instructions due to hallucinations, we incorporate a self-reflection mechanism that enables the VLM to verify the validity of its generated instructions. Additionally, we devise a load balancing strategy to ensure a balanced distribution of editing instructions across various tasks.

\textbf{Edited Image Construction}. We select open-source and closed-source models to create a variety of data construction pipelines tailored to the specifics of different tasks. For instance, in tasks such as subject addition and deletion, which demand high consistency in non-target areas, we employ RAM++\cite{huang2023opensetimagetaggingmultigrained} and SAM2\cite{ravi2024sam2segmentimages} on the original images and subsequently apply LaMa\cite{suvorov2021resolutionrobustlargemaskinpainting} to generate edited images. Our experiments indicate that this methodology results in superior data quality for these tasks compared to models like GPT-4o, Kontext, etc. Details of this comparison are provided in the appendix D. For more general editing tasks, we utilize Step1X-Edit to generate the needed data. In cases where Step1X-Edit performs poorly, such as style modification, we incorporate additional data using OmniConsistency\cite{song2025omniconsistencylearningstyleagnosticconsistency}, TextFlux\cite{xie2025textfluxocrfreeditmodel}, and Kontext. For the challenging construction of complex reasoning and camera movement tasks, we employ GPT-4o and BAGEL. Furthermore, we leverage GPT-4o and Kontext to work with 1024-resolution images with varying aspect ratios
 to generate data that satisfies requirements for high fidelity.

\textbf{Post Quality Enhancement}. In order to guarantee the quality of the data, we implement an extensive framework for data quality evaluation and filtering. We specifically derive aesthetic score, as well as LIQE\cite{zhang2023blindimagequalityassessment} and CLIPIQA\cite{wang2022exploringclipassessinglook} scores for all the generated images, thus facilitating a assessment of image quality and the exclusion of samples that fall below predetermined standards. To ensure the precision of editing, we employ ImgEdit-Judge\cite{ye2025imgeditunifiedimageediting} and Qwen2.5-VL-72B to assess and filter images according to editing instructions, source images, and edited images. For subject-driven generation, we utilize CLIP\cite{hessel2021clipscore} and DINO\cite{caron2021emerging} to assess subject consistency. For style transfer, Qwen2.5-VL-7B evaluates the stylistic match between source and generated images, and applies filters accordingly.

\subsection{Dataset Statistics}
\begin{figure}[ht!]
\centering
\includegraphics[scale=0.1]{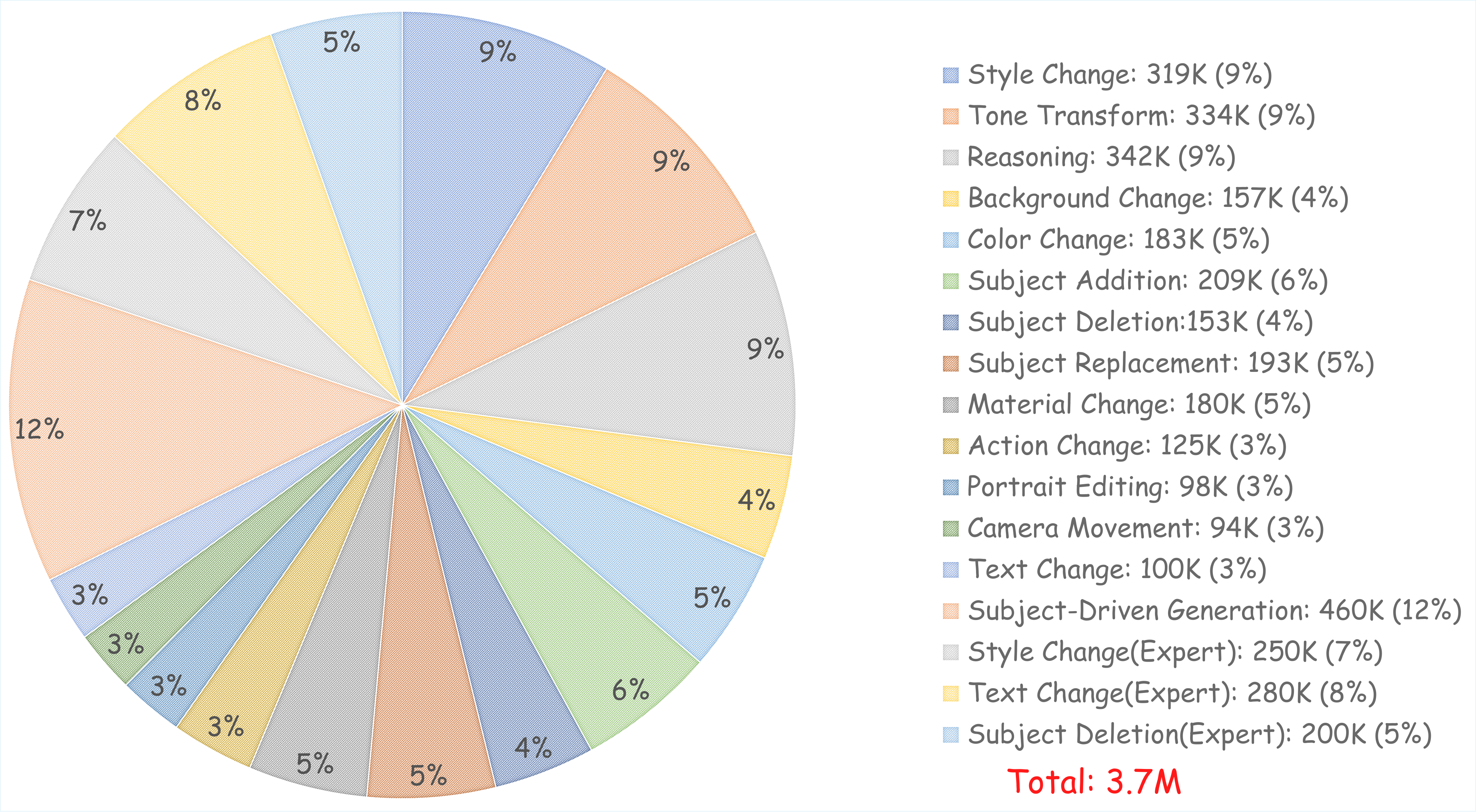}
\caption{X2Edit Dataset Collection Distribution.}
\label{Fig: dataset statistic}
\end{figure}

As shown in Figure~\ref{Fig: dataset statistic}, the X2Edit Dataset comprises 3.7 million pairs of high-quality image across 14 categories. It outperforms existing datasets in terms of task diversity, scale, and resolution, as shown in Table~\ref{Tab: dataset1}. Notably, it includes 342K reasoning editing samples, 94K camera movement samples, and 460K subject-driven generation samples, addressing the scarcity of such data in current open-source datasets. To evaluate editing precision and data quality, we randomly select 1k samples from each dataset and perform comprehensive evaluations using models such as Qwen2.5-VL-72B, ImgEdit-Judge, and GPT-4o. The results indicate that the X2Edit Dataset is competitive against other open-source image editing datasets in terms of VLM evaluations, aesthetics, and overall image quality. Specifically, the high-resolution samples in the X2Edit Dataset surpass almost all existing open-source counterparts. Detailed comparison results are available in the appendix D.

\section{Methodology}

\subsection{Model Overview}
\begin{figure*}[ht!]
	\centering
    \includegraphics[width=1\textwidth]{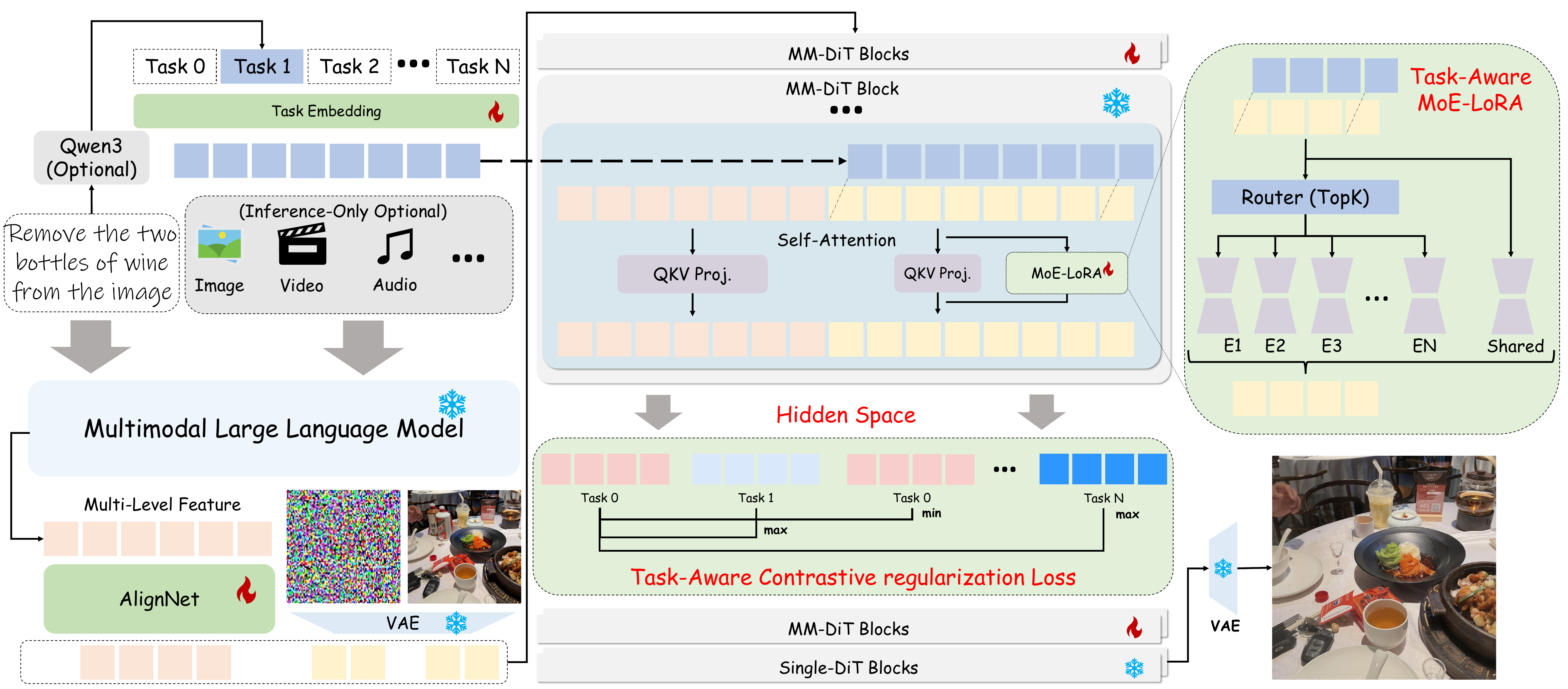}
     \caption{X2Edit consists of an MLLM for editing instruction understanding, a DiT fine-tuned based on FLUX.1, an optional intent perception model, and task embeddings. We introduce a task-aware MoE-LoRA structure and task-aware contrastive learning into the DiT to enhance the unified editing model's ability to perceive different editing tasks.}
  \label{Fig: X2Edit}
\end{figure*}

As shown in Figure~\ref{Fig: X2Edit}, we adopt X2I\cite{ma2025x2iseamlessintegrationmultimodal} as our backbone. The injection of reference image information draws on classic strategies\cite{tan2025ominicontrolminimaluniversalcontrol,zhang2025easycontroladdingefficientflexible,tan2025ominicontrol2efficientconditioningdiffusion,wang2025unicombineunifiedmulticonditionalcombination,mou2025dreamounifiedframeworkimage,ma2023glyphdrawseamlesslyrenderingtext} of concatenating noise. During training, the AlignNet branch within X2I, the task-embedding matrix, and MoE-LoRA parameters are simultaneously updated. We apply task-aware contrastive regularization to the intermediate features of all MMDiT blocks, enforcing a structured hidden space. During inference, we optionally deploy Qwen3 to predict the editing task. We can further leverage X2I's multimodal inputs to enrich the textual editing instructions as shown in the top-right unit of Figure~\ref{Fig:1}.

\subsection{Task-Aware MoE-LoRA}
Image editing tasks span from low-level manipulations to high-level semantic edits. Task heterogeneity can lead to parameter inefficiency when using a single model, as shared parameters internalize interference-prone representations, causing suboptimal specialization and increased parameter numbers. Task-aware MoE addresses this by activating sparse expert sub-networks based on the editing task, allocating capacity precisely where needed.

MoE consists of $N_e$ expert networks, a shared expert network, and a gating network. The gating network outputs an $N_e$-dimensional vector representing the scores of experts, which are subsequently converted into normalized weights via a softmax function. We select the top-$K$ experts with the highest weights for token processing, ensuring computational efficiency through sparse activation. The shared expert processes all tokens to balance knowledge distribution and eliminate redundant storage of shared representations.

We define $h^l \in \mathbb{R}^{b\times n\times d}$ as the intermediate representations at the $l$-th layer of the MMDiT. Let $t_{\text{emb}} \in \mathbb{R}^{N_t \times c}$ represents task embeddings and $y \in \mathbb{R}^{b}$ (where $y_i \in {1, 2, \ldots, N_t}$) indicate the task type corresponding to the $b$ samples of $h^l$, where $N_t$ denotes the number of editing task types. We first extract the task-specific embedding corresponding to $h^l$ from $t_{\text{emb}}$, then reshape and expand it to enable channel-wise concatenation with $h^l$:
\begin{equation}
\label{Eq1}
t_\text{emb}^h = \text{Expand}(\text{Reshape}(t_\text{emb}[y], (b,1,c))).
\end{equation}
Subsequently, we process the concatenation of $t_{\text{emb}}^h$ and $h^l$ through the gating network and the softmax function to obtain the weights of $N_e$ experts:
\begin{equation}
s_{i} = \text{Softmax}_{i} \left( \text{Gate}(\text{Concat}(h^l,t_\text{emb}^h)) \right).
\end{equation}
Finally, we select the top-$K$ experts with the highest weights and aggregate the outputs of these experts with the corresponding gating weights through weighted summation to yield the MoE outputs $q_\text{moe}^{l}$, $k_\text{moe}^{l}$, and $v_\text{moe}^{l}$:
\begin{align}
x_\text{moe}^{l} &= \sum_{i=1}^{N_e} \left( g_{i} \text{Expert}_{x}^i \left( h^{l} \right) \right) \nonumber\\
&+ \text{SharedExpert}_x\left(h^{l}\right),\ \text{for}\ x \in \{ q, k, v \},\\
\label{Eq4}
g_{i} &= \begin{cases} s_{i}, & s_{i} \in \text{Topk} \left( \left\{ s_{j} \middle| 1 \leqslant j \leqslant N_e \right\}, K \right), \\ 0, & \text{otherwise}. \end{cases}
\end{align}
we add $q_\text{moe}^{l}$, $k_\text{moe}^{l}$, and $v_\text{moe}^{l}$ to the corresponding outputs from the linear projector of MMDiT, respectively, and then execute the self-attention mechanism.
\begin{equation}
h_\text{moe}^{l} = \text{Attention}(q_\text{moe}^{l}+q^{l},k_\text{moe}^{l}+k^{l},v_\text{moe}^{l}+v^{l}),
\end{equation}
where $q^{l}$, $k^{l}$, and $v^{l}$ denote the vectors projected from $h^l$ through the linear projector of MMDiT.



\subsection{Task-aware Contrastive Learning}

Current diffusion models primarily rely on regression objective for training, generally lacking explicit regularization of internal representations. To address this, we introduce a task-aware contrastive regularization term during diffusion model training. This explicitly structures the hidden space by dispersing different representations while collapsing similar representations into compact regions, thereby enhancing inter-class separability and intra-class compactness.

Unlike Dispersive Loss which employs a contrastive loss without positives, we adopt a standard contrastive loss. For samples from the same editing task, we enforce region-consistent representations in the hidden space, while promoting maximal dispersion across different tasks to ensure inter-class separability. For the construction of positives and negatives, whereas conventional contrastive learning\cite{chen2020simpleframeworkcontrastivelearning} constructs positives through self-supervision and treats all other images in the batch as negatives, we leverage task labels to construct semantically meaningful samples: all intra-task samples within a batch form positives, while inter-task pairs automatically serve as negatives.

We first perform L2 normalization on $ h^l $ to increase training stability and linear separability in the hidden space, and calculate its corresponding distance matrix $D \in \mathbb{R}^{b\times b}$:
\begin{align}
h_\text{norm}^l &= \text{Normalize}(\text{Reshape}(h^l, (b, n\times d))), \\
D_{ij} &= \|h_\text{norm}^l(i)-h_\text{norm}^l(j)\|_2^2,
\end{align}
where $D_{ij}$ denotes the Squared Euclidean Distance between the representations of the $i$-th sample and $j$-th sample. We then calculate the task mask $M \in \mathbb{R}^{b\times b}$ and apply the InfoNCE\cite{oord2019representationlearningcontrastivepredictive} loss:
\begin{align}
M_{ij}&=\begin{cases}1&\mathrm{if}\ y_i=y_j\ \mathrm{and}\ i\neq j\\0&\mathrm{otherwise}\end{cases},\\
\mathcal{L}_\text{task}&=-\frac{1}{b}\sum_{i=1}^N\log\left(\frac{\sum_{j=1}^N\exp(-\frac{D_{ij}}{\tau})\cdot M_{ij}}{\sum_{k=1}^N\exp(-\frac{D_{ik}}{\tau})}\right),
\end{align}
where $\tau$ is the temperature hyperparameter that controls the model’s discrimination against negatives.

In practice, we add this term to the original diffusion-based objectives $\mathcal{L}_\text{diff}$, the final objective becomes:
\begin{equation}
\mathcal{L} := \mathcal{L}_\text{task} + \lambda \mathcal{L}_\text{diff},
\end{equation}
where $\lambda$ is a hyperparameter that controls the tradeoff between denoising and contrastive learning.

\section{Experiment}

\subsection{Implementation}
Our model configuration uses a LoRA rank of 64, sets the number of expert networks $N_e$ to 12, the number of editing task
types $N_t$ to 15(includes an ``other" editing task), the number of activated experts $K$ to 2, the regularization parameter $\lambda$ to 0.2 and the temperature parameter $\tau$ to 0.5. We train the X2Edit model using 48 H20 GPUs. We conduct 16k training steps at 512-resolution with a micro-batch size of 12, followed by 5k training steps at 1024-resolution with a micro-batch size of 4 on the same GPUs. The random seed for inference on all evaluation is set to 1. 

\subsection{Benchmark Suite}
\textbf{Arbitrary-Instruction Image Editing.}
\textbf{Evaluation datasets} include GEdit-Bench++, ImgEdit-Bench, AnyEdit-Test, and KontextBench\cite{labs2025flux1kontextflowmatching}, each of which includes multiple types of editing tasks. GEdit-Bench++ extends GEdit-Bench by incorporating reasoning and camera movement, expanding the total task count from 11 to 13. For the construction of evaluation data for reasoning and camera movement, we manually select 50 real-world images and generate appropriate instructions in both Chinese and English based on the content of the images. \textbf{Comparative methods} include closed-source models SeedEdit\cite{wang2025seededit} and GPT-4o, as well as open-source models Step1X-Edit, Kontext, Bagel, OmniGen2, ICEdit, Hidream-E, and AnyEdit. All models are evaluated using the same prompts and images. \textbf{Evaluation metrics} employ VIEScore\cite{ku2023viescore} and ImgEdit-Judge Score. VIEScore includes three metrics: Semantic Consistency(SC), Perceptual Quality(PQ), and Overall Score(O). SC measures the alignment of the generated image with prompt, while PQ assesses the visual authenticity and naturalness of the generated image and $\mathrm{O=\sqrt{SC\times PQ}}$. Similar to Step1X-Edit, we use GPT-4o and Qwen2.5-VL-72B to automatically calculate the VIEScore. ImgEdit-Judge Score assesses the degree of instruction following and absence of unintended changes in the generated images.

\textbf{Subject-Driven Generation.}
\textbf{Evaluation dataset} employs classical DreamBench\cite{ruiz2023dreamboothfinetuningtexttoimage}. \textbf{Comparative methods} include recent methods such as X2I, UNO\cite{wu2025lesstomoregeneralizationunlockingcontrollability}, Kontext, Bagel, OmniGen2, OmniGen\cite{xiao2024omnigenunifiedimagegeneration}, ACE++\cite{mao2025aceinstructionbasedimagecreation} and Previous methods. \textbf{Evaluation metrics} include the average DINO and CLIP-I scores between source images and generated images, as well as the average CLIP-T score between generated images and prompts.

\subsection{Quantitative Results}

\begin{table*}[h]
\centering
\footnotesize
\begingroup
\setlength{\tabcolsep}{2pt}
\resizebox{\textwidth}{!}{
\begin{tabular}{cccccccccccccccccc}
\hline
\multirow{2}{*}{Methods} & \multirow{2}{*}{Params}& \multirow{2}{*}{US} & \multicolumn{3}{c}{GEdit-Bench++\_CN} & \multicolumn{3}{c}{GEdit-Bench++\_EN} & \multicolumn{3}{c}{ImgEdit-Bench} & \multicolumn{3}{c}{AnyEdit-Test} & \multicolumn{3}{c}{KontextBench} \\
\cline{4-18}
&  & & IJ & Q\_VIE & G\_VIE & IJ & Q\_VIE & G\_VIE  & IJ & Q\_VIE & G\_VIE & IJ & Q\_VIE & G\_VIE & IJ & Q\_VIE & G\_VIE \\
\hline
GPT-4o & - & 2.911& 9.062 & 7.706 & 7.943 & 9.003 & 7.684 & 7.848 & 8.202 & 7.634 & 7.328 & 8.460 & 7.414 & 6.836 & 8.358 & 7.281 & 7.078 \\
Seededit & -& 2.660 & 8.686 & 6.215 & 6.460 & 8.604 & 6.449 & 6.344 & 8.143 & 6.820 & 6.552 & 8.204 & 6.677 & 5.840 & 8.106 & 6.770 & 5.983 \\\hline
Kontext & 12B & 2.881& -& - & - & 8.408 & 6.170 & 5.712 & 8.149 & 6.087 & 5.258 & 8.110 & 5.419 & 4.900 & 8.095 & 6.250 & 5.718 \\
Bagel & 7B+7B& 2.632 & 8.461 & 6.649 & 5.627 & 8.326 & 6.748 & 5.722 & 7.925 & 6.748 & 6.022 & 7.960 & 6.292 & 5.451 & 7.929 & 6.586 & 5.880 \\
Omnigen2 & 3B+4B& 2.427 & 8.001 & 4.265 & 4.199 & 7.973 & 4.523 & 4.321 & 8.018 & 5.797 & 5.482 & 7.609 & 4.162 & 3.825 & 7.548 & 4.879 & 4.278 \\
Step1X-Edit & 12B& 2.305 & 8.146 & 5.994 & 5.430 & 8.017 & 5.844 & 5.108 & 7.653 & 6.064 & 5.425 & 7.753 & 5.342 & 4.674 & 7.476 & 5.272 & 4.481 \\
Hidream-E & 17B& 2.198 & - & - & - & 7.461 & 5.630 & 4.257 & 7.264 & 6.079 & 4.344 & 6.923 & 5.014 & 3.282 & 7.154 & 4.860 & 3.770 \\\hline
ICEdit & 0.2B& 2.036 & - & - & - & 7.203 & 4.984 & 4.109 & 7.615 & 5.443 & 5.228 & 7.498 & 4.510 & 3.782 & 7.192 & 4.390 & 3.720 \\
AnyEdit & 0.9B& - & - & - & - & 6.841 & 2.608 & 2.242 & 6.784 & 3.991 & 3.448 & 7.078 & 2.884 & 2.528 & 6.452 & 1.959 & 1.760 \\ \hline
X2Edit & 0.9B& 2.432 & 8.313 & 6.354 & 5.639 & 8.334 & 6.158 & 5.550 & 8.025 & 5.987 & 5.402 & 8.095 & 5.578 & 5.147 & 7.606 &5.768 & 5.214 \\ 
\hline
\end{tabular}}
\caption{Objective performance on four benchmark datasets. IJ, Q\_VIE, and G\_VIE refer to the overall score evaluated by ImgEdit-Judge, the VIEScore calculated by Qwen2.5-VL-72B and GPT-4o. US refer to the user study on GEdit-Bench++.}\label{Table-1}
\endgroup
\end{table*}

\begin{table}[t]
\footnotesize
\begingroup
\setlength{\tabcolsep}{5pt}
\resizebox{0.5\textwidth}{!}{
\begin{tabular}{ccccc}
\hline
Methods & Base & DINO & CLIP-I & CLIP-T \\
\hline
Textual Inversion\cite{gal2022imageworthwordpersonalizing} & \multirow{7}{*}{SD} & 0.569 & 0.780 & 0.255 \\
DreamBooth\cite{ruiz2023dreamboothfinetuningtexttoimage} &  & 0.668 & 0.803 & 0.305 \\
BLIP-Diffusion\cite{li2023blipdiffusionpretrainedsubjectrepresentation} &  & 0.594 & 0.779 & 0.300 \\
Subject-Diffusion\cite{ma2024subjectdiffusionopendomainpersonalizedtexttoimage} &  & 0.711 & 0.787 & 0.293 \\
IP-Adapter\cite{ye2023ipadaptertextcompatibleimage} &  & 0.667 & 0.813 & 0.289 \\
KOSMOS-G\cite{pan2024kosmosggeneratingimagescontext} &  & 0.694 & 0.847 & 0.287 \\
\hline
SuTI\cite{chen2023subjectdriventexttoimagegenerationapprenticeship} & Imagen & 0.741 & 0.819 & 0.304 \\
\hline
OmniGen\cite{xiao2024omnigenunifiedimagegeneration} & Phi-3 & - & 0.801 & 0.315 \\
\hline
OmniGen2\cite{wu2025omnigen2explorationadvancedmultimodal} & Qwen-VL & 0.807 & 0.814 & 0.332 \\
\hline
Bagel\cite{deng2025emergingpropertiesunifiedmultimodal} & Qwen & 0.739 & 0.756 & 0.332 \\
\hline
UNIC-Adapter\cite{duan2024unicadapterunifiedimageinstructionadapter} & SD3 & 0.816 & 0.841 & 0.306 \\
\hline
IP-Adapter\cite{ye2023ipadaptertextcompatibleimage} & \multirow{7}{*}{FLUX.1} & 0.768 & 0.803 & 0.322 \\
OminiControl\cite{tan2025ominicontrolminimaluniversalcontrol} &  & 0.740 & 0.768 & 0.329 \\
Kontext\cite{labs2025flux1kontextflowmatching} &  & 0.822 & 0.839 & 0.322 \\
UNO\cite{wu2025lesstomoregeneralizationunlockingcontrollability} &  & 0.764 & 0.806 & 0.319 \\
ACE++\cite{mao2025aceinstructionbasedimagecreation} &  & 0.760 & 0.787 & 0.319 \\
X2I\cite{ma2025x2iseamlessintegrationmultimodal} &  & 0.817 & 0.826 & 0.304 \\ \hline
X2Edit & FLUX.1 & 0.822 & 0.826 & 0.326 \\
\hline
\end{tabular}}
\caption{Performance of different methods on DreamBench.}\label{Table-2}
\endgroup
\end{table}

\textbf{Arbitrary-Instruction Image Editing}. Table~\ref{Table-1} shows the performance of different methods on four evaluation datasets. In the evaluation on GEdit-Bench++, X2Edit achieves scores of 8.313, 6.354, and 5.639 for the Chinese sub-dataset, and scores of 8.334, 6.158, and 5.55 for the English sub-dataset. These results indicate that X2Edit is on par with other mainstream open-source methods such as Bagel and Kontext, while surpassing Step1X-Edit and OmniGen2 across multiple metrics. Additionally, it significantly outperforms ICEdit, Hidream-E and AnyEdit across all metrics. X2Edit exhibits robust cross-lingual capability, appendix G demonstrates more language tests. On the ImgEdit-Bench and AnyEdit-Test benchmarks, X2Edit achieves performance comparable to that of GEdit-Bench++. It is close to Kontext and Bagel in most metrics, outperforms OmniGen2 and Step1X-Edit in the majority of indicators, and significantly surpasses ICEdit, Hidream-E, and AnyEdit across all metrics.
The KontextBench is somewhat different from the other three benchmarks, especially under the evaluation systems of Qwen2.5-VL-72B and GPT-4o. X2Edit shows a significant gap compared to the two open-source methods, Kontext and Bagel. Nevertheless, X2Edit's support for style transfer and subject-driven generation also results in a significant performance gap compared to the other five comparative methods. KontextBench comprises 1,026 unique image-prompt pairs derived from 108 base images, including personal photos, CC-licensed art, public domain images, and AI-generated content. Compared with other benchmarks, it is relatively complex. Additional subjective comparisons are provided in the appendix H.

\textbf{Subject-Driven Generation}. X2Edit has a slight advantage in the DINO score on DreamBench, while it holds a middle position in the CLIP-I score. 
In terms of the CLIP-T metric, X2Edit slightly lags behind OmniControl, OmniGen2, and Bagel. However, it outperforms these three methods in the other two image fidelity metrics. Overall, X2Edit still demonstrates a relatively competitive result.

\textbf{Summary}. Overall, X2Edit holds its own among top-tier performers. While it still has room to close the gap with closed-source models, it shows a clear edge over fully trained models like Step1X-Edit, OmniGen2, and HiDream-E. It runs neck and neck with Bagel and Kontext, and it markedly surpasses lightweight fine-tuning models such as ICEdit; On DreamBench, X2Edit exhibits enhanced subject-driven generation surpassing most competitors. We additionally assess out-of-domain generalization across complicated editing tasks. Please see the appendix F for details.


\subsection{Plug-and-Play}
\begin{table}[h!]
\centering
\footnotesize
\begin{tabular}{cccc}
\hline
Models & Steps & GC & GE \\
\hline
FLUX.1-Schnell & 4 & 8.254 & 8.085 \\
Shuttle-3-Diffusion & 4 & 8.401 & 8.185 \\
FLUX.1-Krea-dev & 28 & 8.412  & 8.367  \\
PixelWave & 28 & 8.458 & 8.324 \\\hline
FLUX.1-Turbo-Alpha & 8 & 8.258 & 8.154 \\
FLUX.1-dev-LoRA-AntiBlur & 28 & 8.414 & 8.104 \\
FLUX-Midjourney-Mix2-LoRA & 28 & 8.399 & 8.237 \\
FLUX-Super-Realism-LoRA & 28 & 8.247 & 8.115 \\
FLUX-Chatgpt-Ghibli-LoRA & 28 & 7.954 & 7.885 \\\hline
X2Edit & 28 & 8.313 & 8.334 \\
\hline
\end{tabular}
\caption{X2Edit transfers seamlessly to Flux.1-based modules. GC and GE refer to the Chinese and English subset of GEdit-Bench++, we utilize ImgEdit-Judge for evaluation.}\label{plug-and-play}
\end{table}

As shown in Table~\ref{plug-and-play}, we validate the plug-and-play capability of X2Edit on two types of modules with high frequency of use in the Flux.1 community. We provide more subjective results in the appendix E. 

\textbf{FLUX.1 Dev Variants}. FLUX.1-Schnell and Shuttle-3-Diffusion are 4-step accelerated models. PixelWave can generate images of multiple artistic styles. FLUX.1-Krea-dev\cite{flux1kreadev2025} offers strong performance with highly distinctive aesthetics and exceptional realism. The experimental results indicate that through flexible plug-and-play capability, we can improve inference speed while maintaining comparable performance.
\textbf{FLUX.1 Dev LoRA Ecosystem}. FLUX.1-Turbo-Alpha supports 8-step inference. FLUX.1-dev-LoRA-AntiBlur enhances depth of field and clarity, while FLUX-Midjourney-Mix2-LoRA emulates MidJourney v6's distinctive aesthetic. FLUX-Super-Realism-LoRA and FLUX-Chatgpt-Ghibli-LoRA specialize in stylization. The experimental results demonstrate that X2Edit seamlessly adapts to the FLUX.1 ecosystem. 


\subsection{Ablation Study}

\begin{table}[t]
\footnotesize
\begingroup
\setlength{\tabcolsep}{3pt}
\begin{tabular}{ccccccc}
\hline
Methods & Loss & Layers & Experts & Rank & GC & GE \\
\hline
LoRA & - & - & - & 512 & 7.834 & 7.649 \\
MoE-LoRA & - & - & 6 & 128 & 7.943 & 7.751 \\
MoE-LoRA w/TA & - & - & 6 & 128 & 8.087 & 7.985 \\
MoE-LoRA w/TA & - & - & 12 & 64 & 8.161 & 8.084 \\
X2Edit & cosine & 4 & 12 & 64 & 8.253 & 8.113 \\
X2Edit & L2 & 4 & 12 & 64 & 8.289 & 8.120 \\\hline
X2Edit & L2 & all & 12 & 64 & 8.313 & 8.334 \\
\hline
\end{tabular}
\caption{Ablation results on GEdit-Bench++.}\label{ablation}
\endgroup
\end{table}

We design a comprehensive suite of ablation experiments on both English and Chinese sub-datasets of GEdit-Bench++ with ImgEdit-Judge, systematically dissecting the contributions of MoE routing, task-awareness(TA), expert cardinality, and the contrastive regularization.
First, a vanilla single-rank LoRA baseline confirms its limited general-purpose editing capability. Incorporating task priors into our MoE-LoRA yields consistent positive gains on all metrics in Table~\ref{ablation}. Notably, increasing the number of experts while proportionally reducing the LoRA rank, which preserves the total parameter budget, delivers further improvements. This evidences that expert granularity trumps capacity: task-specific low-rank sub-spaces retain ample discriminative power, and the routing diversity from more experts offsets any capacity loss from rank compression, thus affirming the ``narrow-yet-numerous'' expert approach for diverse editing tasks.
Furthermore, we conduct three ablation studies specifically on the contrastive regularization loss. Firstly, we use the conventional cosine similarity and select representations in the fourth layer of MMDiT. The experiments show that the inclusion of regularization loss further enhance the overall performance. Similar to dispersive loss, we explore the impact of different similarity metrics and the position of the regularization layer on performance. Ultimately, we choose squared L2 distance as the similarity metric and calculate the regularization loss using the representations in all 19 layers.

\subsection{User Study}
 We recruit four participants to assess the editing outputs of X2Edit and other comparative methods across 1.3k editing pairs from GEdit-bench++. The evaluation primarily focuses on two aspects: instruction following and image fidelity. The instruction following score measures how effectively the model understands and executes given instructions, while the image fidelity score assesses the overall quality and consistency of edited images. Both scores range from 0 to 2: poor = 0, fair = 1, good = 2. We finally sum these two scores to calculate an overall score. The third column of Table~\ref{Table-1} represents the mean overall score across all participants. We select the GEdit-Bench++ benchmark and calculate the average for models that support both Chinese and English. Given the subpar overall subjective performance of AnyEdit, we do not conduct human annotations. The overall conclusion indicates that our method indeed exhibits a certain level of competitiveness among numerous approaches. For more detailed chart experiments, please refer to the appendix B.

\section{Conclusion and Limitations}

In this paper, we construct and release the X2Edit Dataset, a large-scale, high-quality corpus containing 3.7 million samples across 14 distinct editing tasks. In addition, we design X2Edit, a lightweight and plug-and-play editing model that introduces a novel framework. Extensive experiments validate that the X2Edit Dataset surpasses existing open-source datasets in both quality and task diversity. Concurrently, the X2Edit model demonstrates performance that is competitive with, and in some cases superior to, state-of-the-art models on multiple benchmarks. Its exceptional plug-and-play capability allows for seamless integration into the broader FLUX.1 ecosystem. Our method's limitations include weaker performance on Non-English text change, which is constrained by the capabilities of the base model. Further failure-case analyses are provided in the appendix I. Looking ahead, we will delve deeper into this direction.


\clearpage

\bibliography{aaai2026}

\clearpage 

\appendix

\section*{Appendix}

\quad \textbf{Appendix~\ref{Acceleration of Training}} details the calculation during training.

\textbf{Appendix~\ref{User Study}} analyses the subjective scores evaluated by human of each sub-task in GEdit-Bench++.

\textbf{Appendix~\ref{GPT4o Detailed Review}} analyses the overall VIEScore evaluated by GPT-4o on GEdit-Bench++.

\textbf{Appendix~\ref{More Details About the X2Edit Dataset}} details the construction of X2Edit Dataset.

\textbf{Appendix~\ref{Plug-and-Play}} shows the editing results of MoE-LoRA as a pluggable module.

\textbf{Appendix~\ref{Zero-Shot Task}} demonstrates the performance of X2Edit on out-of-domain editing tasks.

\textbf{Appendix~\ref{X2Edit Generation}} demonstrates the results generated by X2Edit in multiple languages and tasks.

\textbf{Appendix~\ref{Visual Comparison}} demonstrates visual comparisons of X2Edit and other comparative methods.

\textbf{Appendix~\ref{Limitations}} discusses the limitations of X2Edit and future work.

\section{Acceleration of Training}
\label{Acceleration of Training}

Through hierarchical asynchronous communication, we achieve overlapping of computation and communication while promptly reclaiming intermediate variables to reduce GPU memory footprint, thereby increasing the micro-batch size. Simultaneously, we employ a gather operation to aggregate sample representations across all distributed devices, addressing issues of insufficient negatives and gradient computation bias. Specifically, for the hidden state of each noise map, we obtain all samples A within the global batch size via all-gather. We then compute the Euclidean distance between the micro-batch samples B on the current GPU and samples A. Following the InfoNCE computation logic, the numerator is derived from the Euclidean distance of positive samples, while the denominator is calculated from the Euclidean distances of both positive and negative samples.


\section{User Study}
\label{User Study}

As shown in Tables~\ref{tab:user-study-1}-\ref{tab:user-study-3},  ``Following" and ``Fidelity" refer to instruction following and image fidelity respectively. ``Average" refers to the average score for these 13 tasks. Although X2Edit lags behind the two unified models Bagel and Kontext, it outperforms several advanced open-source methods such as Step1X-Edit and OmniGen2 in terms of average scores across all tasks. However, it demonstrates relatively weaker performance in text change; the text change editing evaluation encompasses Chinese modifications. Given that the base model inherently lacks the capability for Chinese text generation, it is challenging to rely solely on LoRA fine-tuning to bestow this ability.

\section{GPT4o Detailed Review}
\label{GPT4o Detailed Review}

\begin{figure*}[!h]
\centering
\includegraphics[width=1\textwidth]{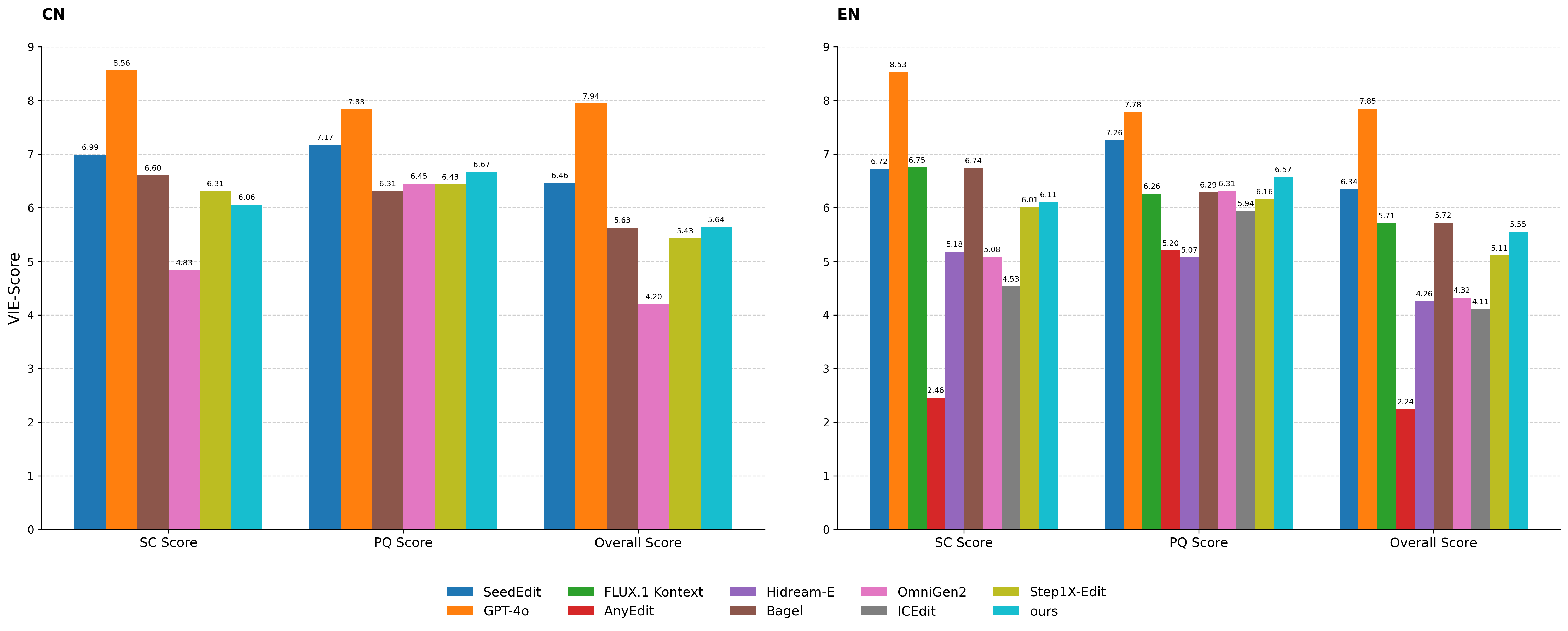}
\caption{PQ score, SC score and overall VIEScore evaluated by GPT-4o on GEdit-Bench++.}\label{GEdit++_bar}
\end{figure*}

As illustrated in Figure~\ref{GEdit++_bar}, X2Edit demonstrates significant advantages over multiple open-source methods in both editing accuracy and visual quality. On the English subset, X2Edit outperforms Step1X-Edit across all three metrics and surpasses all open-source methods in PQ score, while achieving overall scores on par with Kontext and Bagel. On the Chinese subset, X2Edit similarly exceeds all open-source methods in PQ score and outperforms Step1X-Edit in overall score, while matching the overall performance of Bagel. This indicates its superior image fidelity capability and reaffirms its consistent and balanced performance across both English and Chinese instructions.

\begin{figure*}[!h]
\centering
\subfigure[]{
    \includegraphics[width=0.47\textwidth]{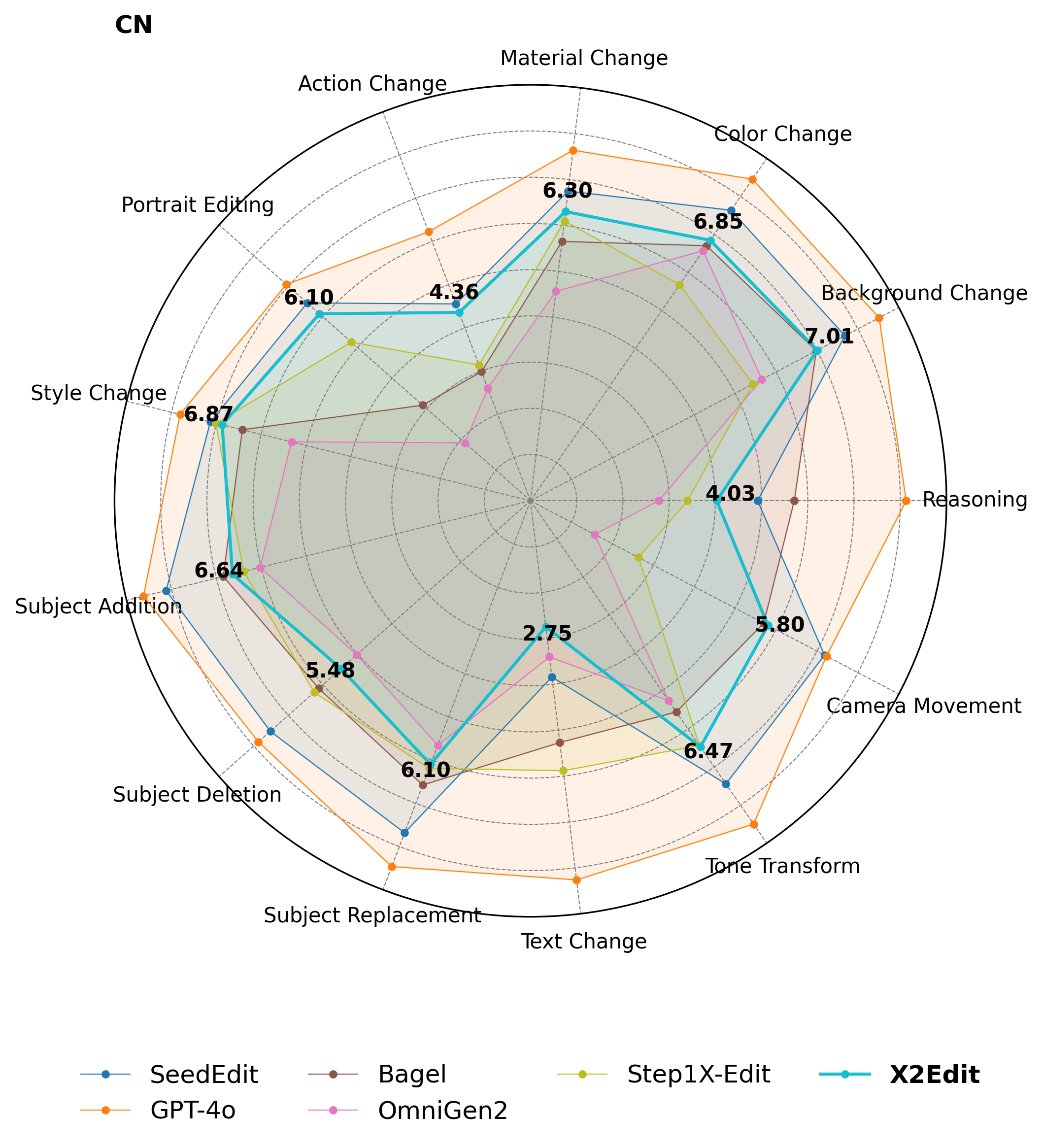}
    \label{GEdit++_cn_radar}
}
\subfigure[]{
    \includegraphics[width=0.48\textwidth]{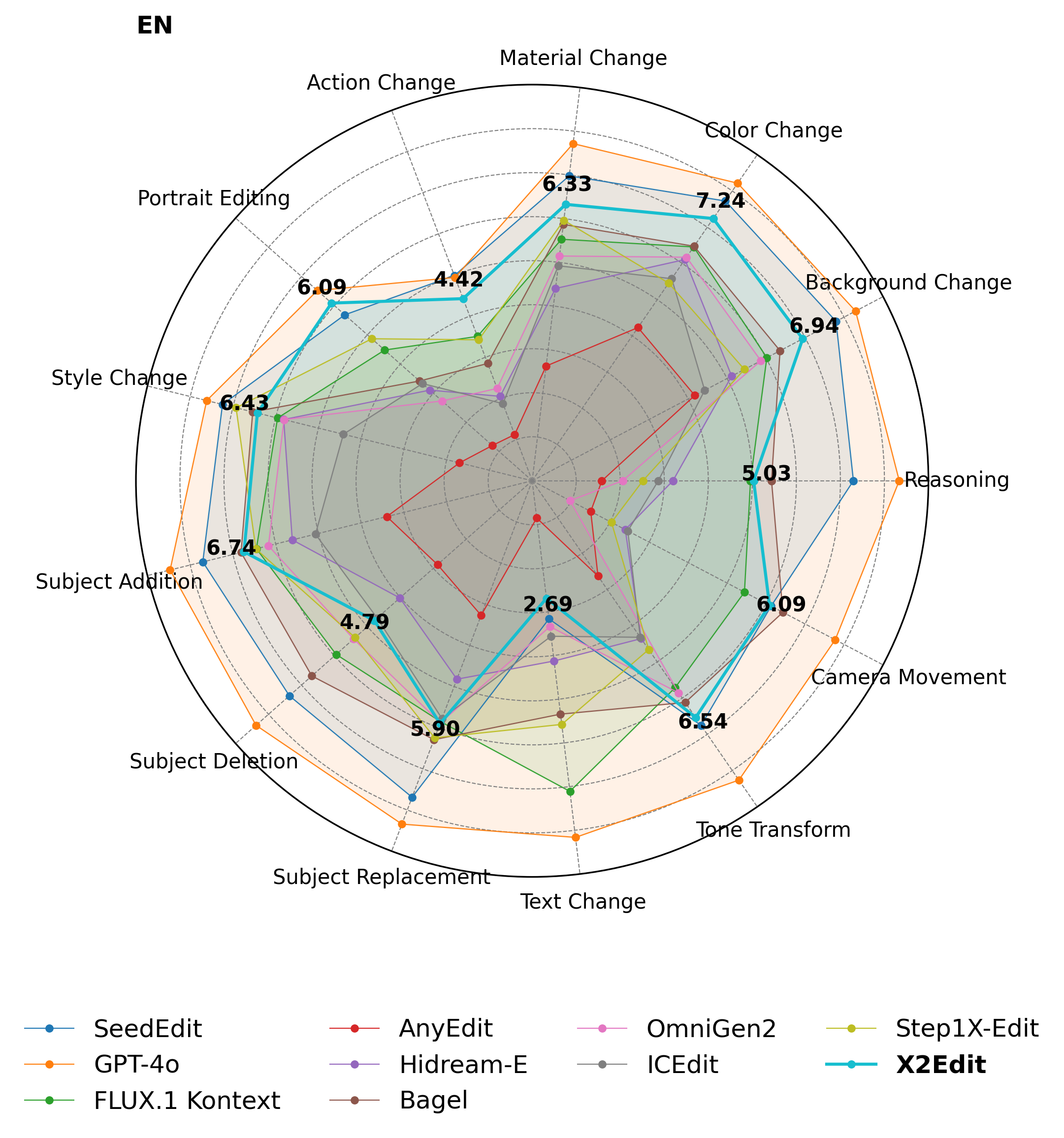}
    \label{GEdit++_en_radar}
}
\caption{Overall VIEScore evaluated by GPT-4o of each sub-task in GEdit-Bench++.}\label{GEdit++_radar}
\end{figure*}

Figure~\ref{GEdit++_radar} demonstrate the VIEScore performance of X2Edit and comparative methods in each sub-task on GEdit-Bench++. In general, X2Edit outperforms existing advanced open-source methods in multiple editing tasks, including tasks that involve camera movement and background change, while achieving leading performance in reasoning tasks among all comparative methods. It also achieves very similar evaluation scores for Chinese and English instructions.






\begin{table*}[!h]
\centering
\footnotesize
\begingroup
\setlength{\tabcolsep}{2pt}
\begin{tabular}{c|ccc|ccc|ccc|ccc|ccc}
\hline
\multirow{2}{*}{Method} & \multicolumn{3}{c|}{Reasoning} & \multicolumn{3}{c|}{Camera Movement} & \multicolumn{3}{c|}{Backeground Change} & \multicolumn{3}{c|}{Color Change} & \multicolumn{3}{c}{Material Change} \\ \cline{2-16} 
 & Following & Fidelity & Sum & Following & Fidelity & Sum & Following & Fidelity & Sum & Following & Fidelity & Sum & Following & Fidelity & Sum \\ \hline
GPT-4o & 1.63 & 1.35 & 2.98 & 1.87 & 1.25 & 3.12 & 2.00 & 0.96 & 2.96 & 2.00 & 1.11 & 3.11 & 1.77 & 1.16 & 2.93 \\
SeedEdit & 1.41 & 1.44 & 2.85 & 1.43 & 1.25 & 2.68 & 1.80 & 1.03 & 2.83 & 1.69 & 1.06 & 2.75 & 1.64 & 1.04 & 2.68 \\
Hidream-E & 0.48 & 0.85 & 1.33 & 0.38 & 0.75 & 1.13 & 1.51 & 1.51 & 3.02 & 1.33 & 1.89 & 3.22 & 0.84 & 1.17 & 2.01 \\
Step1X-Edit & 0.70 & 0.54 & 1.24 & 0.44 & 0.45 & 0.89 & 1.30 & 1.23 & 2.53 & 1.34 & 1.14 & 2.48 & 1.49 & 0.97 & 2.46 \\
ICEdit & 0.44 & 0.91 & 1.35 & 0.14 & 0.85 & 0.99 & 0.92 & 1.47 & 2.39 & 1.40 & 1.38 & 2.78 & 1.08 & 1.05 & 2.13 \\
Kontext & 0.59 & 1.18 & 1.77 & 0.60 & 1.25 & 1.85 & 1.61 & 1.95 & 3.56 & 1.50 & 1.78 & 3.28 & 1.18 & 1.48 & 2.66 \\
Bagel & 0.74 & 1.24 & 1.98 & 0.35 & 1.40 & 1.75 & 1.91 & 0.73 & 2.64 & 1.46 & 1.58 & 3.04 & 1.67 & 0.82 & 2.49 \\
OmniGen2 & 0.57 & 0.58 & 1.15 & 0.11 & 0.11 & 0.22 & 1.42 & 1.70 & 3.12 & 1.27 & 1.74 & 3.01 & 0.75 & 1.57 & 2.32 \\
\hline
X2Edit & 0.73 & 1.12 & 1.85 & 0.93 & 1.05 & 1.98 & 1.83 & 1.13 & 2.96 & 1.72 & 1.04 & 2.76 & 1.24 & 1.21 & 2.45 \\ \hline
\end{tabular}
\caption{Detailed subjective scores evaluated by human of each sub-task in GEdit-Bench++.}\label{tab:user-study-1}
\endgroup
\end{table*}

\begin{table*}[!h]
\centering
\footnotesize
\begingroup
\setlength{\tabcolsep}{2pt}
\begin{tabular}{c|ccc|ccc|ccc|ccc|ccc}
\hline
\multirow{2}{*}{Method} & \multicolumn{3}{c|}{Action Change} & \multicolumn{3}{c|}{Portrait Editing} & \multicolumn{3}{c|}{Style Change} & \multicolumn{3}{c|}{Subject Addition} & \multicolumn{3}{c}{Subject Deletion} \\ \cline{2-16} 
 & Following & Fidelity & Sum & Following & Fidelity & Sum & Following & Fidelity & Sum & Following & Fidelity & Sum & Following & Fidelity & Sum \\ \hline
GPT-4o & 1.94 & 0.92 & 2.86 & 1.97 & 0.90 & 2.87 & 1.66 & 1.19 & 2.85 & 1.91 & 0.99 & 2.90 & 1.83 & 0.92 & 2.75 \\
SeedEdit & 1.56 & 0.96 & 2.52 & 1.72 & 1.18 & 2.90 & 1.56 & 1.24 & 2.80 & 1.70 & 1.14 & 2.84 & 1.64 & 1.01 & 2.65 \\
Hidream-E & 1.03 & 0.86 & 1.89 & 0.76 & 1.21 & 1.97 & 1.36 & 0.92 & 2.28 & 1.35 & 1.52 & 2.87 & 1.18 & 1.25 & 2.43 \\
Step1X-Edit & 1.24 & 0.98 & 2.22 & 1.38 & 1.22 & 2.60 & 1.34 & 1.11 & 2.45 & 1.48 & 1.38 & 2.86 & 1.31 & 1.27 & 2.58 \\
ICEdit & 0.85 & 1.05 & 1.90 & 0.56 & 1.40 & 1.96 & 0.97 & 1.07 & 2.04 & 1.15 & 1.10 & 2.25 & 1.02 & 1.37 & 2.39 \\
Kontext & 0.93 & 1.73 & 2.66 & 0.79 & 1.85 & 2.64 & 1.05 & 1.87 & 2.92 & 1.67 & 1.88 & 3.55 & 1.47 & 1.98 & 3.45 \\
Bagel & 1.41 & 1.36 & 2.77 & 1.72 & 0.95 & 2.67 & 1.64 & 1.04 & 2.68 & 1.65 & 1.18 & 2.83 & 1.49 & 1.69 & 3.18 \\
OmniGen2 & 0.81 & 1.51 & 2.32 & 0.55 & 1.73 & 2.28 & 1.04 & 1.64 & 2.68 & 1.32 & 1.69 & 3.01 & 1.23 & 1.89 & 3.12 \\
\hline
X2Edit & 1.48 & 0.99 & 2.47 & 1.27 & 1.02 & 2.29 & 1.44 & 0.93 & 2.37 & 1.49 & 1.08 & 2.57 & 1.24 & 1.21 & 2.45 \\ \hline
\end{tabular}
\caption{Detailed subjective scores evaluated by human of each sub-task in GEdit-Bench++.}\label{tab:user-study-2}
\endgroup
\end{table*} 

\begin{table*}[!h]
\centering
\footnotesize
\begingroup
\setlength{\tabcolsep}{5pt}
\begin{tabular}{c|ccc|ccc|ccc|ccc}
\hline
\multirow{2}{*}{Method} & \multicolumn{3}{c|}{Subject Replacement} & \multicolumn{3}{c|}{Text Change} & \multicolumn{3}{c|}{Tone Transform} & \multicolumn{3}{c}{\textbf{Average}} \\ \cline{2-13} 
 & Following & Fidelity & Sum & Following & Fidelity & Sum & Following & Fidelity & Sum & Following & Fidelity & Sum \\ \hline
GPT-4o & 1.83 & 0.95 & 2.78 & 1.72 & 1.01 & 2.73 & 2.00 & 1.00 & 3.00 & 1.90 & 1.01 & 2.911 \\
SeedEdit & 1.68 & 1.01 & 2.69 & 0.54 & 1.00 & 1.54 & 1.83 & 1.02 & 2.85 & 1.59 & 1.07 & 2.660 \\
Hidream-E & 1.22 & 1.00 & 2.22 & 0.77 & 1.16 & 1.93 & 0.78 & 1.50 & 2.28 & 0.96 & 1.24 & 2.198 \\
Step1X-Edit & 1.50 & 1.07 & 2.57 & 1.42 & 1.22 & 2.64 & 1.22 & 1.22 & 2.44 & 1.12 & 1.18 & 2.305 \\
ICEdit & 1.12 & 1.00 & 2.12 & 0.74 & 1.16 & 1.90 & 0.85 & 1.42 & 2.27 & 0.83 & 1.21 & 2.036 \\
Kontext & 1.31 & 1.43 & 2.74 & 1.46 & 1.94 & 3.40 & 1.39 & 1.58 & 2.97 & 1.11 & 1.77 & 2.881 \\
Bagel & 1.62 & 0.93 & 2.55 & 0.78 & 1.57 & 2.35 & 1.73 & 1.55 & 3.28 & 1.41 & 1.22 & 2.632 \\
OmniGen2 & 1.09 & 1.43 & 2.52 & 0.59 & 1.66 & 2.25 & 1.58 & 1.97 & 3.55 & 0.74 & 1.69 & 2.427 \\
\hline
X2Edit & 1.41 & 1.20 & 2.61 & 0.72 & 1.19 & 1.91 & 1.63 & 1.31 & 2.94 & 1.31 & 1.12 & 2.432 \\ \hline
\end{tabular}
\caption{Detailed subjective scores evaluated by human of each sub-task in GEdit-Bench++.}\label{tab:user-study-3}
\endgroup
\end{table*}

\section{More Details About the X2Edit Dataset}
\label{More Details About the X2Edit Dataset}

\subsection{Diverse Editing Instruction Generation.}
To address the limitations in instruction diversity and consistency during the instruction generation process, we use prompt engineering techniques to carefully design prompts that drive VLM, as shown below:

\textit{If you are a talented photo artist, refer to the content of the given input image and give reasonable prompts for image editing instructions.}

\textbf{\textit{Note:}}
\begin{itemize}
    \item \textit{Give appropriate image editing instructions according to the content of the image, without outputting the intermediate process, and select the appropriate 10 instruction types. For example, if there is no core character in the image, there is no need for the character editing instruction type.}
    \item \textit{Each type corresponds to 1 specific instruction, and try to cover each instruction type.}
    \item \textit{Please refer to the context example for the type and output format of the editing instruction. Only refer to its format, and the diversity of the output content is greater. For example, the color change is not limited to: becoming clear, adjusting the color, adding filters, but also includes night to day, repairing old photos, weather defogging, time to night, etc.; the styles include Ghibli, illustration, oil painting, wool felt, miniature landscape, two-dimensional, pixel art, clay, Pixar, line drawing art, 3D cartoon, abstract, ink, watercolor, animation, abstract art, cartoon, etc.}
\end{itemize}

We include detailed examples and clear instructions in the prompts to improve the accuracy, balance, and rationality of VLM-constructed editing instructions. Additionally, we also provide contextual examples of each editing task for the VLM, in the hope that it could imitate the format to generate similar editing instructions. The contextual examples are shown as follows:

\begin{itemize}
    \item \textit{``Background Change'': ``Change the background from indoor to outdoor''}
    \item \textit{``Color Change'': ``Change the red clothes to green''}
    \item \textit{``Material Change'': ``Change the metal cup to ceramic''}
    \item \textit{``Action Change'': ``Change the person to a waving pose''}
    \item \textit{``Portrait Editing'': ``Make the portrait more handsome and the skin better''}
    \item \textit{``Style Change'': ``Change to Ghibli style''}
    \item \textit{``Subject Addition'': ``Add a puppy to the right side of the picture''}
    \item \textit{``Subject Deletion'': ``Remove the trash can in the lower left corner of the photo''}
    \item \textit{``Subject Replacement'': ``Change the kitten to a puppy''}
    \item \textit{``Text Change'': ``Change the title from `Hello' to `Hi'"}
    \item \textit{``Tone Tranform'': ``Change the cold tone to warm tone''}
\end{itemize}

We use the above workflow to drive the generation of editing instruction data for various editing tasks. Note that for workflows that generate certain specific editing tasks, such as text change and subject-driven generation, their editing instruction generation process does not adopt the above general template. Their specific workflow information can be found in Section~\ref{Edited Image Data Construction}.

\subsection{Category Balanced Strategy}
Existing datasets typically suffer from the problem of imbalanced editing categories, leading to poor performance of trained models on certain tasks. To mitigate this issue, we perform category balancing during the instruction generation phase. Specifically, by caching instructions and counting the frequency of their types, we can dynamically adjust weights based on the proportion of different instruction types in the current cache to select a certain number of different types of instructions, thereby achieving the goal of balancing instruction types.

\begin{figure*}[htbp]
\centering
\includegraphics[scale=0.25]{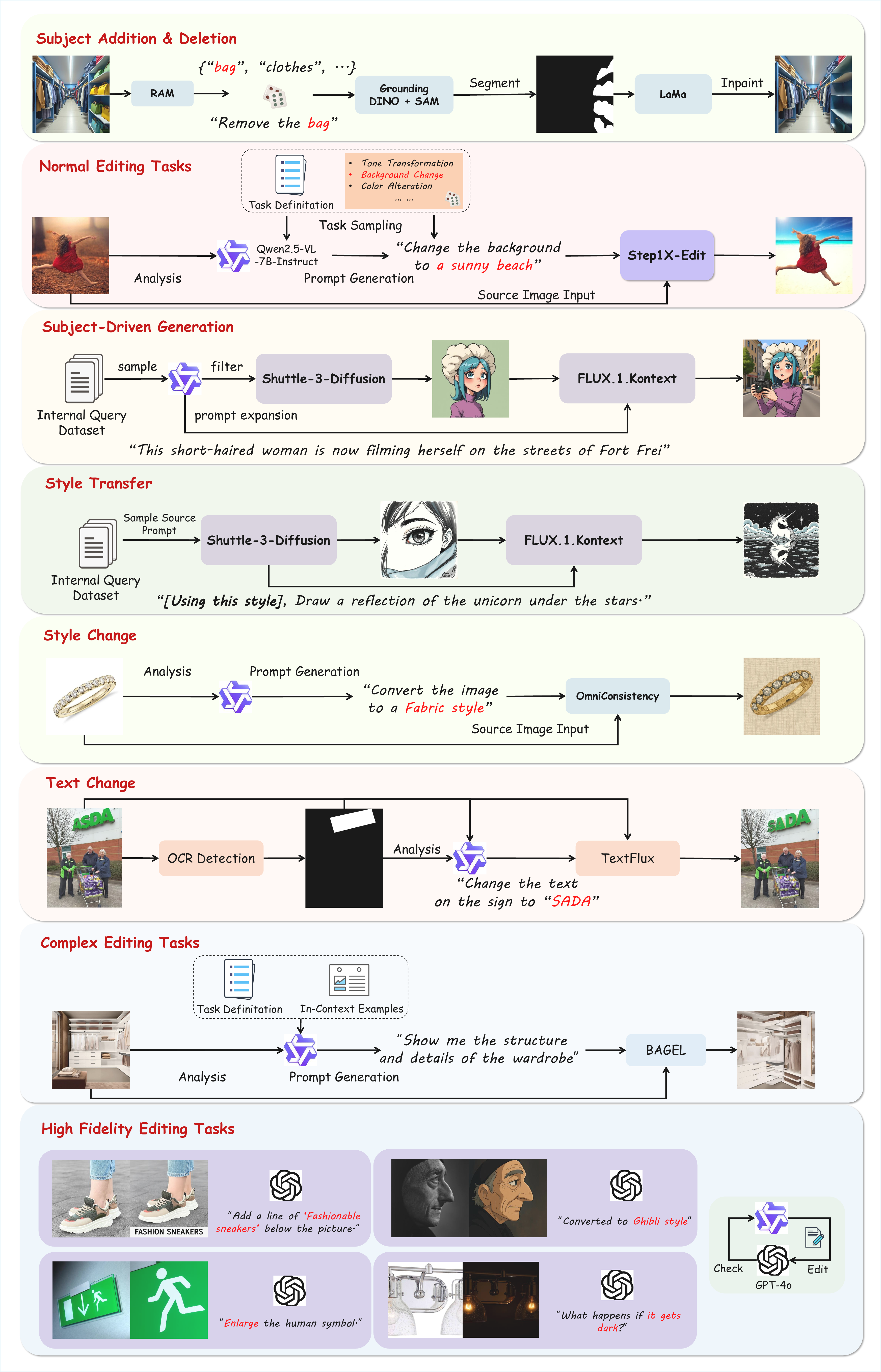}
\caption{Detailed dataset construction process of X2Edit Dataset.}
\label{Fig: dataset detail}
\end{figure*}

\begin{table*}[t]
\centering
\footnotesize
\begingroup
\setlength{\tabcolsep}{2.3pt}
\begin{tabular}{l|ccc|ccc|ccc|cc|cccc}
\hline
\multirow{2}{*}{Datasets} & \multicolumn{3}{c|}{Qwen2.5-VL-72B} & \multicolumn{3}{c|}{ImgEdit-Judge} & \multicolumn{3}{c|}{GPT-4o} & \multicolumn{2}{c|}{Aesthetics} & \multicolumn{4}{c}{Image Quality} \\ \cline{2-16} 
 & S\_1 & S\_2 & Overall & S\_1 & S\_2 & Overall & SC & PQ & Overall & Src & Tar & Src\_LIQE & Tar\_LIQE & Src\_clip & Tar\_clip \\ \hline
AnyEdit & 3.628 & 3.566 & 7.194 & 3.611 & 4.180 & 7.791 & 6.738 & 6.411 & 6.025 & 4.746 & 4.549 & 4.523 & 4.259 & 0.715 & 0.688 \\
HQ-Edit & 3.897 & 3.739 & 7.636 & 3.917 & 4.590 & 8.507 & 6.971 & 8.017 & 7.176 & 5.834 & 5.727 & 4.568 & 4.510 & 0.740 & 0.733 \\
Ultra-Edit & 3.782 & 3.486 & 7.268 & 4.108 & 4.410 & 8.518 & 7.473 & 6.258 & 6.587 & 4.221 & 4.265 & 3.978 & 3.847 & 0.655 & 0.648 \\
SEED-Data-Edit & 3.731 & 3.859 & 7.590 & 3.730 & 4.594 & 8.324 & 6.081 & 6.161 & 5.221 & 4.048 & 4.056 & 3.207 & 3.205 & 0.593 & 0.594 \\
Img-Edit & 3.526 & 3.809 & 7.335 & 3.616 & 4.509 & 8.125 & 5.533 & 7.318 & 5.415 & 4.925 & 4.905 & 4.219 & 4.209 & 0.681 & 0.677 \\
OmniEdit & 4.212 & 4.159 & 8.371 & 4.331 & 4.697 & 9.028 & 6.927 & 8.030 & 7.019 & 5.030 & 5.059 & 4.205 & 4.111 & 0.697 & 0.686 \\ \hline
X2Edit(512) & 3.849 & 3.924 & 7.773 & 4.504 & 4.668 & 9.171 & 6.724 & 6.123 & 5.869 & 4.600 & 5.015 & 3.574 & 3.840 & 0.600 & 0.635 \\
X2Edit(1024) & 3.833 & 4.248 & 8.081 & 4.938 & 4.992 & 9.930 & 6.722 & 7.629 & 6.326 & 5.847 & 5.716 & 4.538 & 4.540 & 0.727 & 0.735 \\ \hline
\end{tabular}
\caption{Overall dataset quality comparison between X2Edit Dataset and other datasets. S\_1 and S\_2 refer to the score of instruction following and absence of unintended changes in the ImgEdit-Judge benchmark.}
\label{Tab: dataset2}
\endgroup
\end{table*}

\begin{table}[t]
\centering
\footnotesize
\begingroup
\setlength{\tabcolsep}{2.5pt}
\begin{tabular}{ccccc}
\hline
Task & X2Edit & Step1X-Edit & Bagel & Kontext \\ \hline
Subject Deletation & 7.197 & 6.3 & 6.98 & 6.402 \\
Style Change & 6.957 & 6.821 & 6.895 & 6.382 \\
Text Change (CN) & 6.323 & 6.147 & 4.783 & - \\
Text Change (EN) & 6.911 & 6.116 & 6.537 & 5.404 \\ \hline
\end{tabular}
\caption{A comparison of data quality for four tasks constructed using our pipeline versus those built using other editing models. We use the same source images and editing instructions and employ GPT-4o to evaluate editing accuracy in order to compare the strengths and weaknesses of different data construction pipelines.}
\label{Tab: dataset3}
\endgroup
\end{table}

\subsection{VLM Self-Reflection}
Since VLMs are prone to hallucinations and may generate editing instructions that do not match image content or are unreasonable, even though we have already added many constraints and examples in the prompts. Therefore, we use VLMs to evaluate the editing instructions they generate themselves, which we call self-reflection. Our prompt is as follows:

\textit{You are a professional digital artist. You need to evaluate the effectiveness of AI-generated images according to a given rule. All input images are AI-generated. All people in the images are also AI-generated, so you don't need to worry about privacy issues.}

\textit{\textbf{Rules}: Two images will be provided: the first is the original AI-generated image, and the second is the first image after editing. The goal is to evaluate how successfully the editing instructions were executed in the second image. Please note that sometimes the two images may look exactly the same, which is due to failed image editing. }

\textit{\textbf{Scoring rules:}}
\begin{itemize}
    \item \textit{The first score is scored from 0 to 10: according to the degree of editing success, it is scored from 0 to 10 (0 means that the edited image does not comply with the editing instructions at all, and 10 means that the edited image complies with the editing instructions completely).}
    \item \textit{The second score is scored from 0 to 10: to evaluate the degree of over-editing, to evaluate whether only the editing instructions are modified, while the rest of the image is not modified (0 means that the edited image is completely different from the original image or the image quality is very poor, and 10 means that the edited image can be considered a minimal but effective edit).}
\end{itemize}
\textit{Put the scores in a list, and output score = [score1, score2], where 'score1' evaluates the success of the edit and 'score2' evaluates the degree of over-editing.}

\textit{\textbf{Editing instructions:}} $<$instruction$>$

We filter out samples below the threshold based on the VLM self-reflection results, thereby improving the accuracy and reasonableness of editing instructions.

\subsection{Edited Image Data Construction}
\label{Edited Image Data Construction}
This section will elaborate on the pipeline implementation details for various editing tasks in the X2Edit Dataset. Figure~\ref{Fig: dataset detail} shows the illustrations of the main specific pipelines in our edited image data construction.

\textbf{Subject Addition \& Deletion. }To obtain prompts, we first use the RAM model to identify object categories (such as ``dog", ``car", etc.) and filter prompts with length $>$ 15 to ensure concise instructions. We then input the prompts into Grounding DINO to generate object bounding boxes and map them to original image dimensions. Subsequently, we use the SAM model to generate precise segmentation masks based on bounding boxes and discard invalid masks with area ratios $<$ 2\% or $>$ 35\%. Finally, we use the LAMA model to perform image inpainting based on masks to delete target objects.

\textbf{Normal Editing Tasks. } We use Qwen2.5-VL-7B to generate editing instructions based on sampled source images according to preset editing task types, driving Step1X-Edit to execute edits on source images.

\textbf{Subject-Driven Generation. }We first sample image descriptions from internal query datasets and input them into Qwen to determine if there are foreground entity keywords for filtering. we then use sampled image descriptions with shuttle-3-diffusion to generate reference images, adding solid-color simple backgrounds. Subsequently, we drive Qwen to expand reference image descriptions, such as changing backgrounds, scenes, and actions, as prompts for Kontext. Finally, we input reference images and expanded prompts into Kontext to generate images.

\textbf{Style Transfer. }We sample two image descriptions from internal query datasets, one for inputting into shuttle-3-diffusion to generate reference images, the other for driving Kontext to generate final images based on the reference image's style.

\textbf{Style Change. }We use Qwen2.5-VL-7B to randomly select a style [Style] based on sampled source images, forming the prompt ``Convert the image to a [Style]'', and input this prompt into OmniConsistency to change the style of the source image.

\textbf{Text Change. }We use OCR detection models to pre-extract text content and region coordinates from source images. During data generation, we parse text region coordinates, generate masks, and select the text region with the largest area as the editing target, ensuring the ratio $\ge$ 1\%. We then use Qwen2.5-7B-VL to generate instructions based on text content, changing original text to target text while constraining the text length difference before and after editing to $\le$ 3. Subsequently, we concatenate the original image and mask as input and use TextFlux to generate edited images based on instructions. Finally, we use Qwen2.5-7B-VL to detect text in edited images, confirming target text has been modified.

\textbf{Complex Editing Tasks. }Process is largely the same as \textbf{Normal Editing Tasks}, except requiring Qwen2.5-VL-7B to generate editing instructions for complex editing tasks based on sampled source images and using Bagel to generate edited images.

\textbf{High Fidelity Editing Tasks. }Process is largely the same as Normal Editing Tasks, using higher resolution source images, and employing GPT-4o and Kontext to generate data for various editing tasks mentioned above.

\subsection{More Examples of X2Edit Dataset}
\begin{figure*}[htbp]
\centering
\includegraphics[scale=0.16]{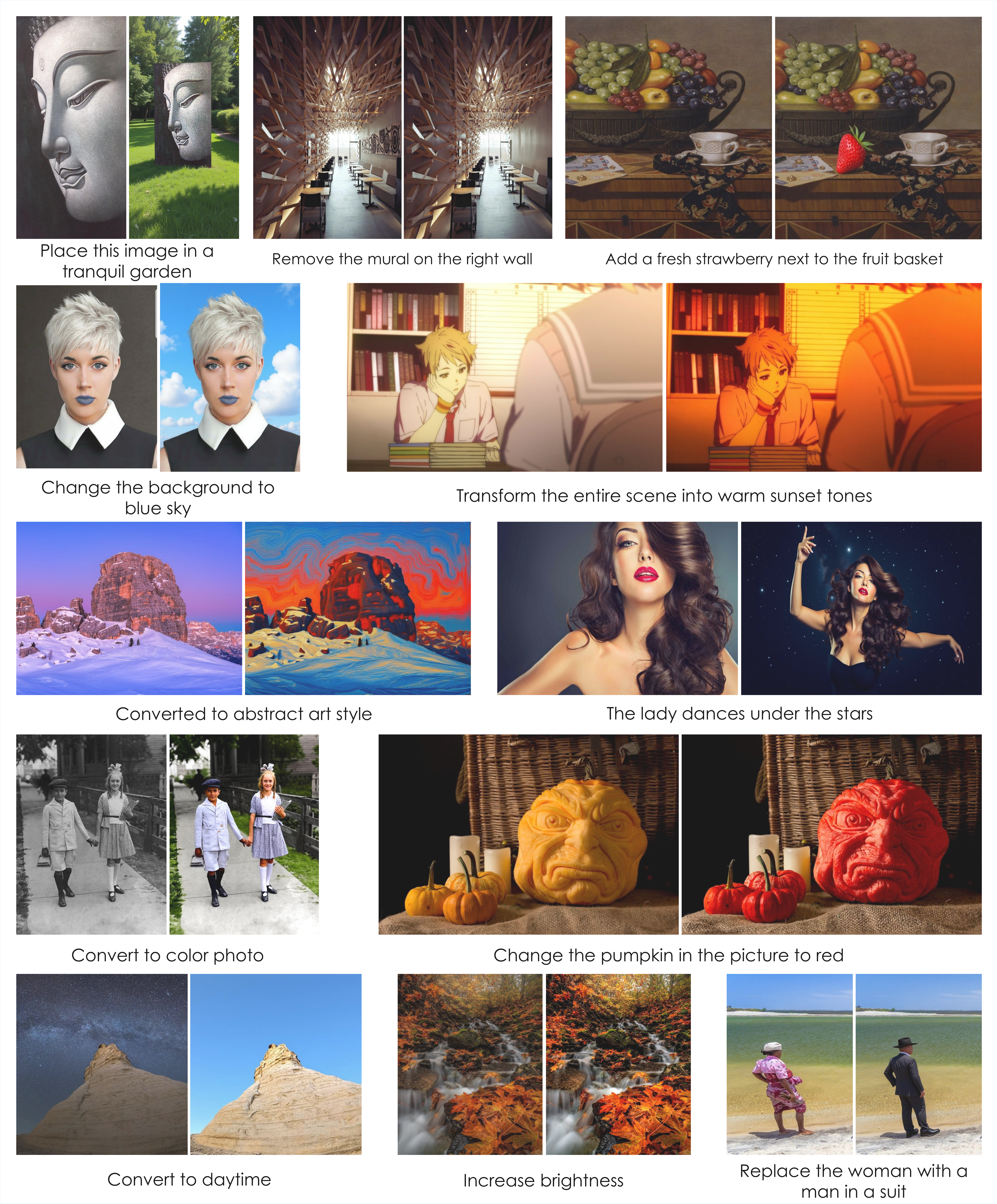}
\caption{High-quality samples from our X2Edit Dataset.}
\label{Fig: dataset case}
\end{figure*}

See Figure~\ref{Fig: dataset case} for more data examples with various editing types in X2Edit Dataset.

\subsection{More experimental results on X2Edit Dataset}
Tables~\ref{Tab: dataset2} and \ref{Tab: dataset3} show more comparison and analysis results on the X2Edit Dataset. Table~\ref{Tab: dataset2} shows a detailed comparison of X2Edit with other open source datasets in multiple dimensions including ImgEdit-Judge, Qwen2.5-VL-72B evaluation (using the ImgEdit-Judge benchmark), GPT-4o evaluation (using the VIE Score benchmark), aesthetic score, and overall image quality. Note that we use the Qwen2.5-VL-72B evaluation based on the VIEScore benchmark when screening data, so for the sake of fairness, we use the ImgEdit-Judge benchmark for comparison here. The results show that our dataset is very competitive among current open-source datasets in all indicators, especially the quality of high-resolution datasets is better than almost all existing datasets. Table~\ref{Tab: dataset3} demonstrates the superiority of our process for constructing subject addition \& deletion task data, in which we use the GPT-4o evaluation based on the VIEScore benchmark.

\section{Plug-and-Play}
\label{Plug-and-Play}

\begin{figure*}[h]
\centering
\includegraphics[width=0.95\textwidth]{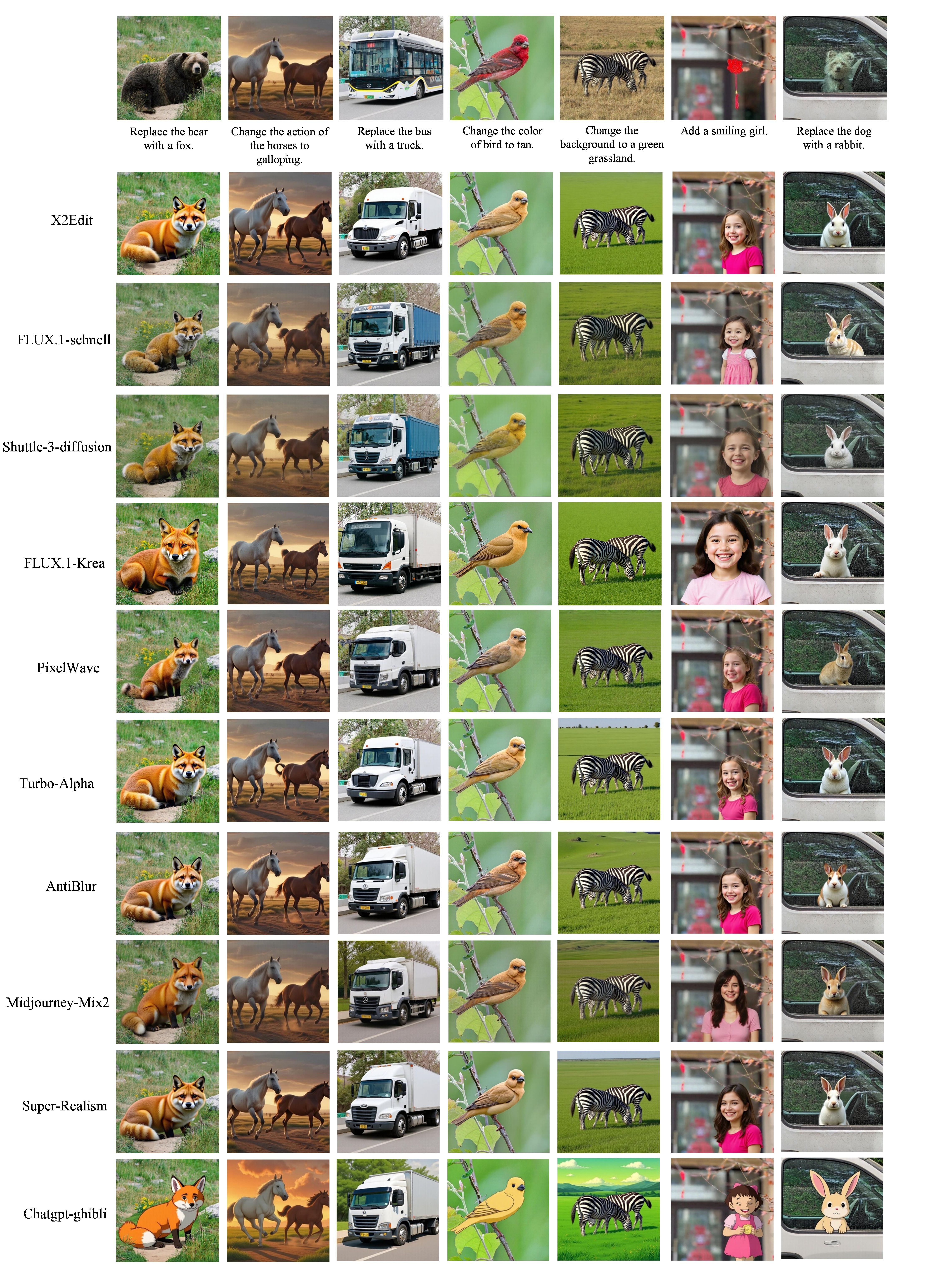}
\caption{Editing results of integrating X2Edit with multiple models and LoRA modules}\label{comparison 2}
\end{figure*}

As illustrated in Figure~\ref{comparison 2} , we demonstrate image editing results of integrating the MoE-LoRA module with multiple accelerated models and diverse LoRA modules. The integration with accelerated models achieves accelerated inference while preserving editing capabilities, while the integration with other LoRA modules maintain both editing performance and the distinctive characteristics of these LoRA modules.

\section{Zero-Shot Task}
\label{Zero-Shot Task}

\begin{figure*}[h]
\centering
\includegraphics[width=0.97\textwidth]{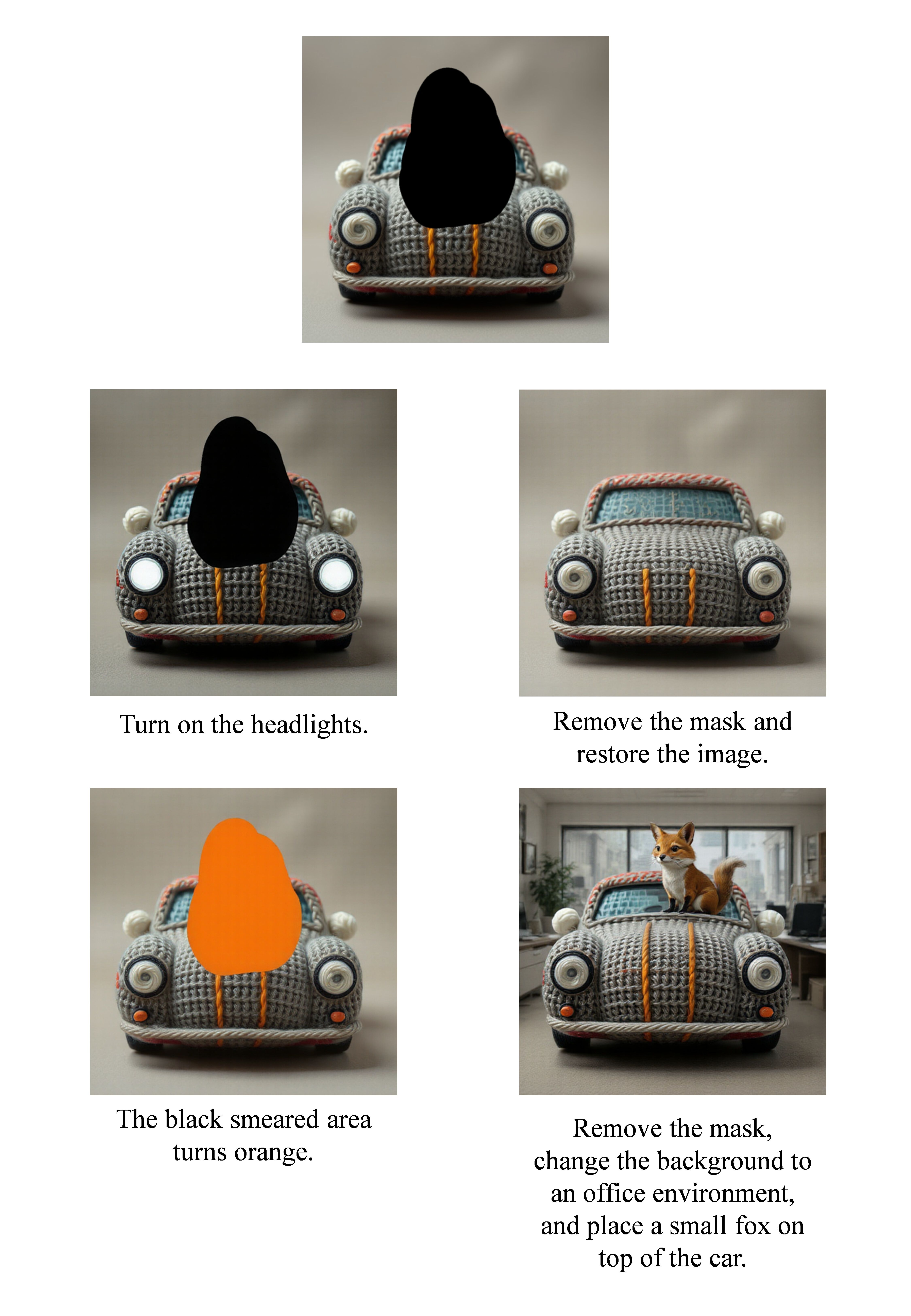}
\caption{Samples of X2Edit generation on zero-shot task.}\label{zero-shot}
\end{figure*}

In X2Edit, in addition to the 14 predefined editing tasks, we introduce an ``other" category. When an input editing instruction does not correspond to any of these 14 tasks, we classify it as the ``other" type. This enables the model to handle editing tasks not present in the training set, thereby enhancing its generalization capabilities. As shown in the Figure~\ref{zero-shot}, although image inpainting is not among these 14 editing tasks, X2Edit still effectively handles this task while also competently executing complex multi-task editing instructions that include image inpainting operations.

\section{X2Edit Generation}
\label{X2Edit Generation}

\begin{figure*}[h]
\centering
\includegraphics[width=0.97\textwidth]{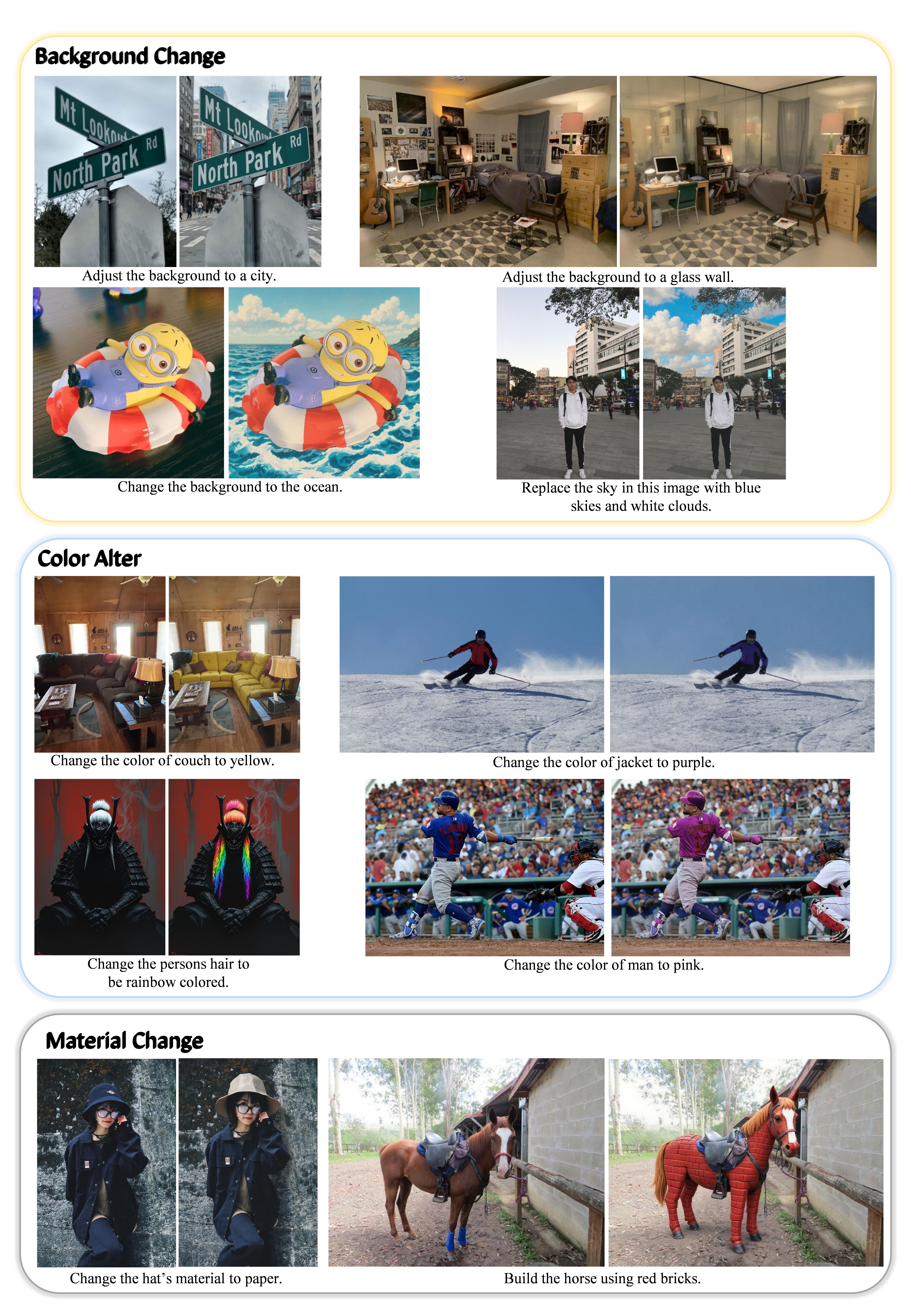}
\caption{Samples of X2Edit generation.}\label{generation 1}
\end{figure*}

\begin{figure*}[h]
\centering
\includegraphics[width=0.97\textwidth]{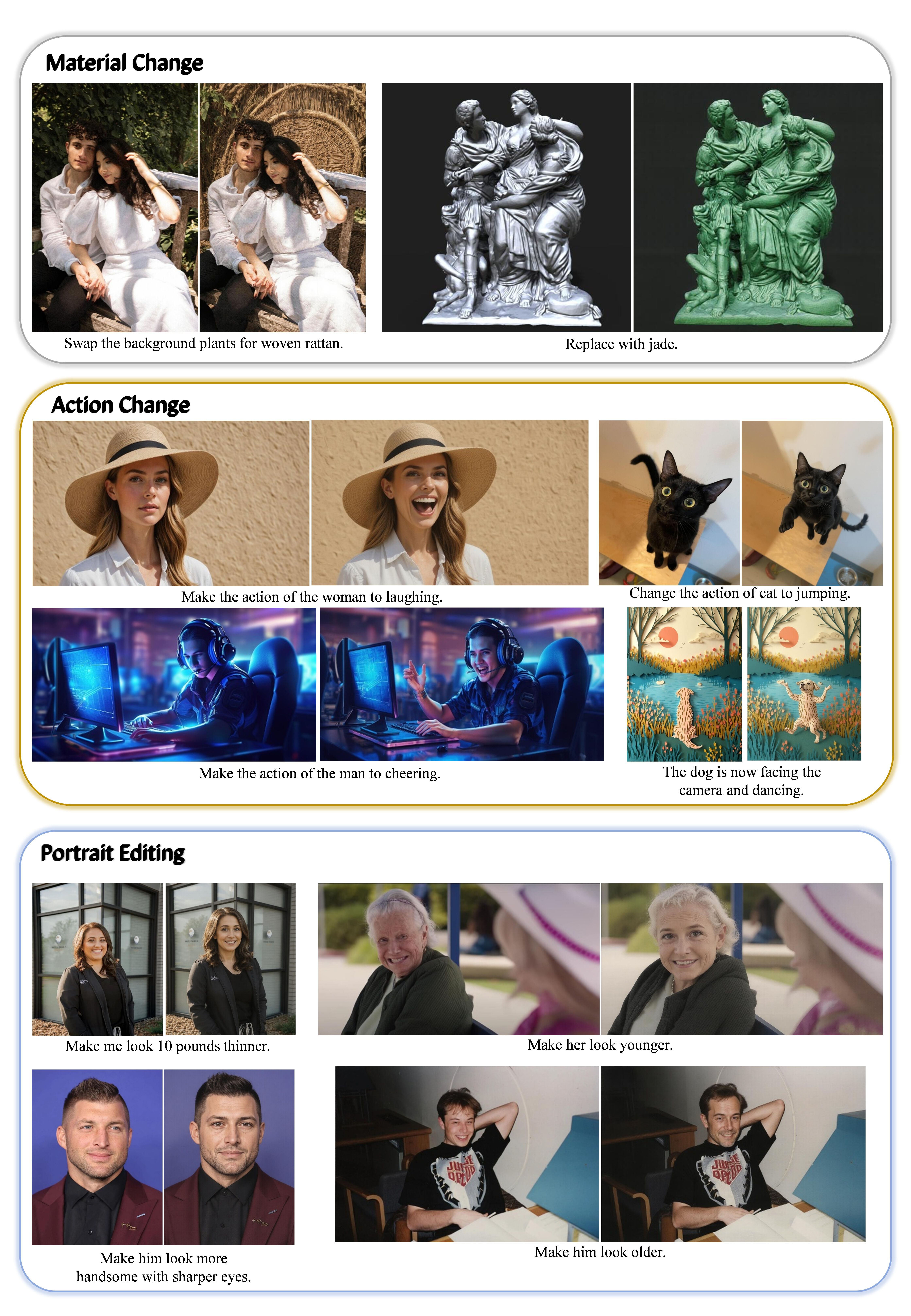}
\caption{Samples of X2Edit generation.}\label{generation 2}
\end{figure*}

\begin{figure*}[h]
\centering
\includegraphics[width=0.97\textwidth]{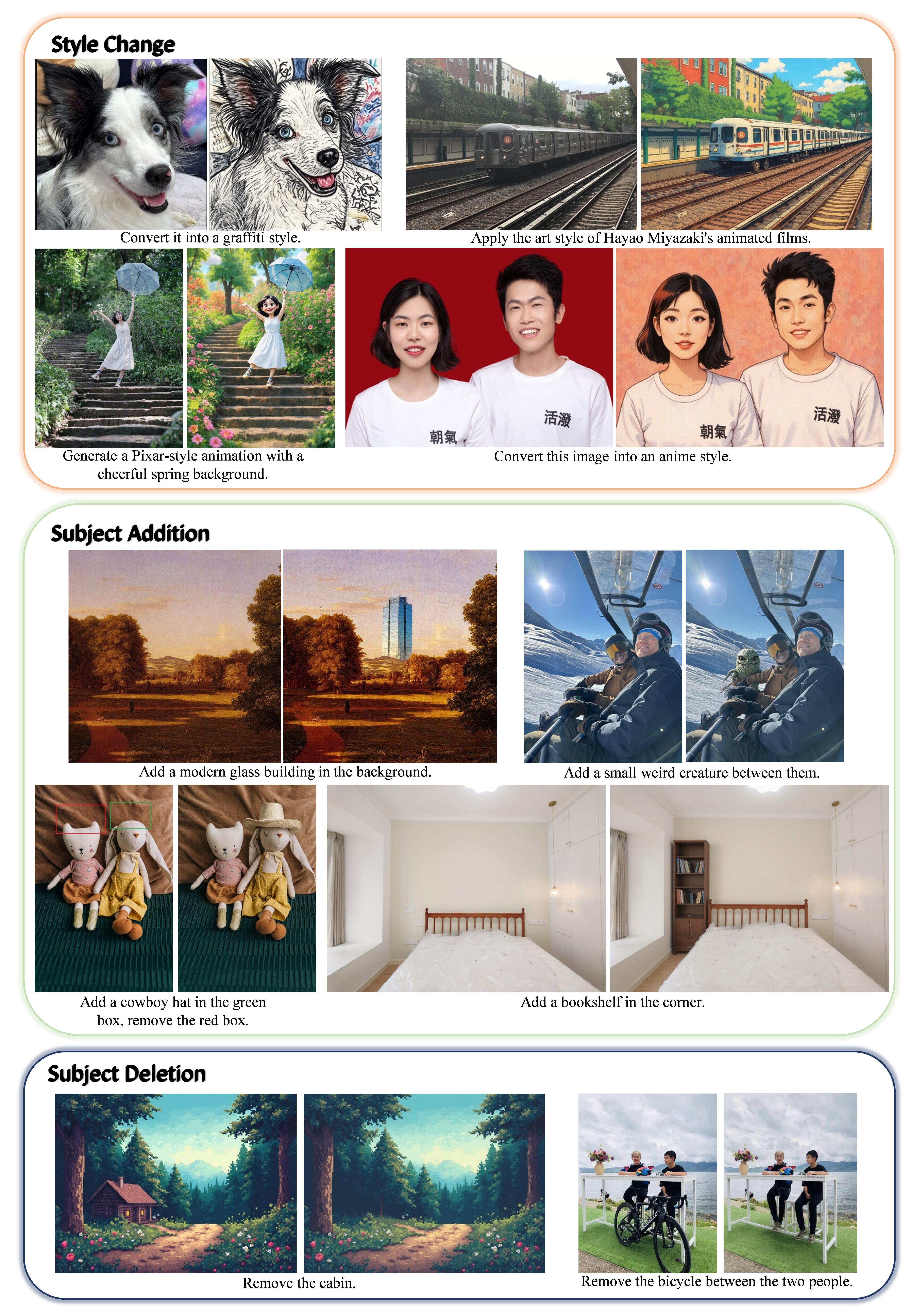}
\caption{Samples of X2Edit generation.}\label{generation 3}
\end{figure*}

\begin{figure*}[h]
\centering
\includegraphics[width=0.97\textwidth]{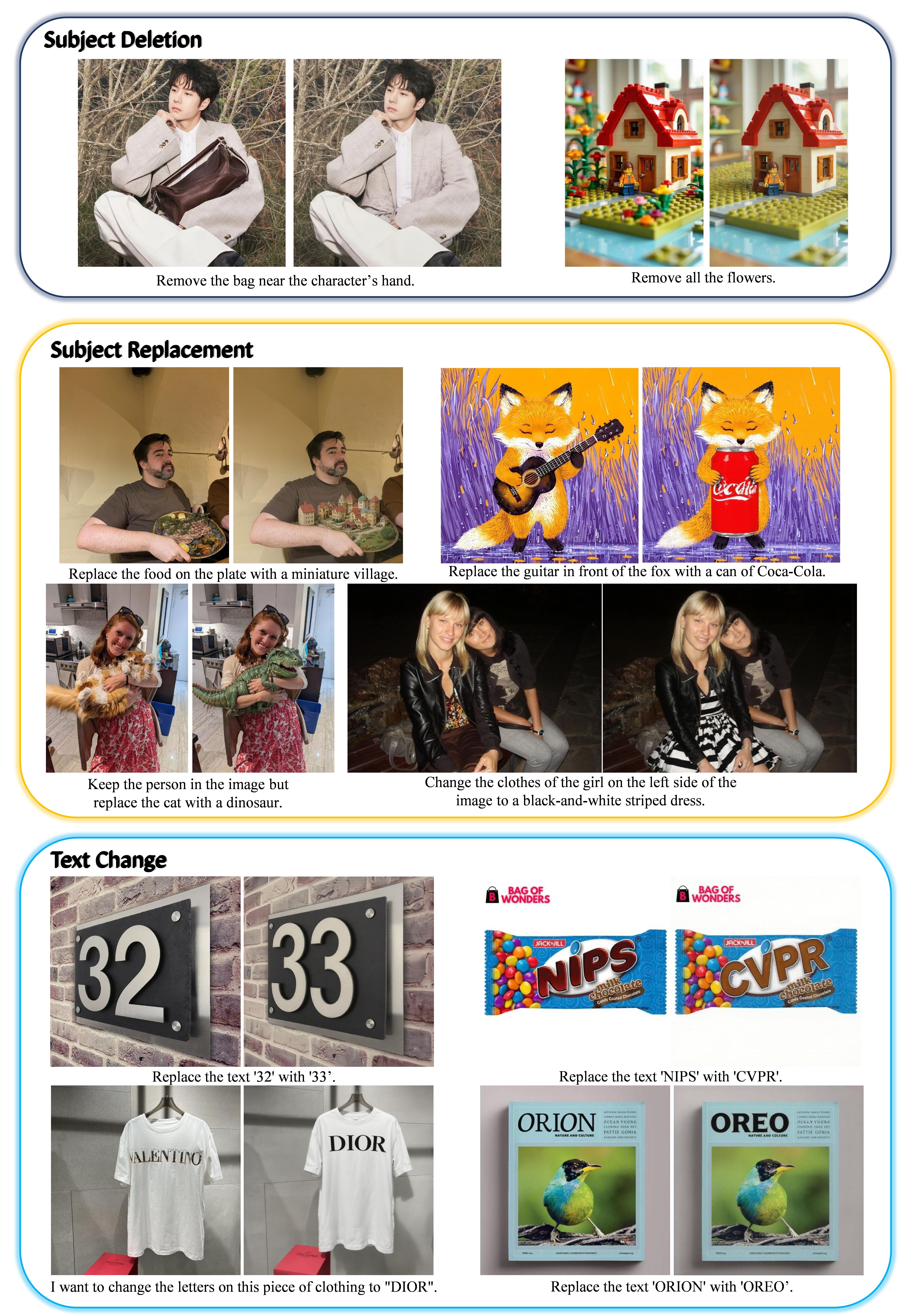}
\caption{Samples of X2Edit generation.}\label{generation 4}
\end{figure*}

\begin{figure*}[h]
\centering
\includegraphics[width=0.97\textwidth]{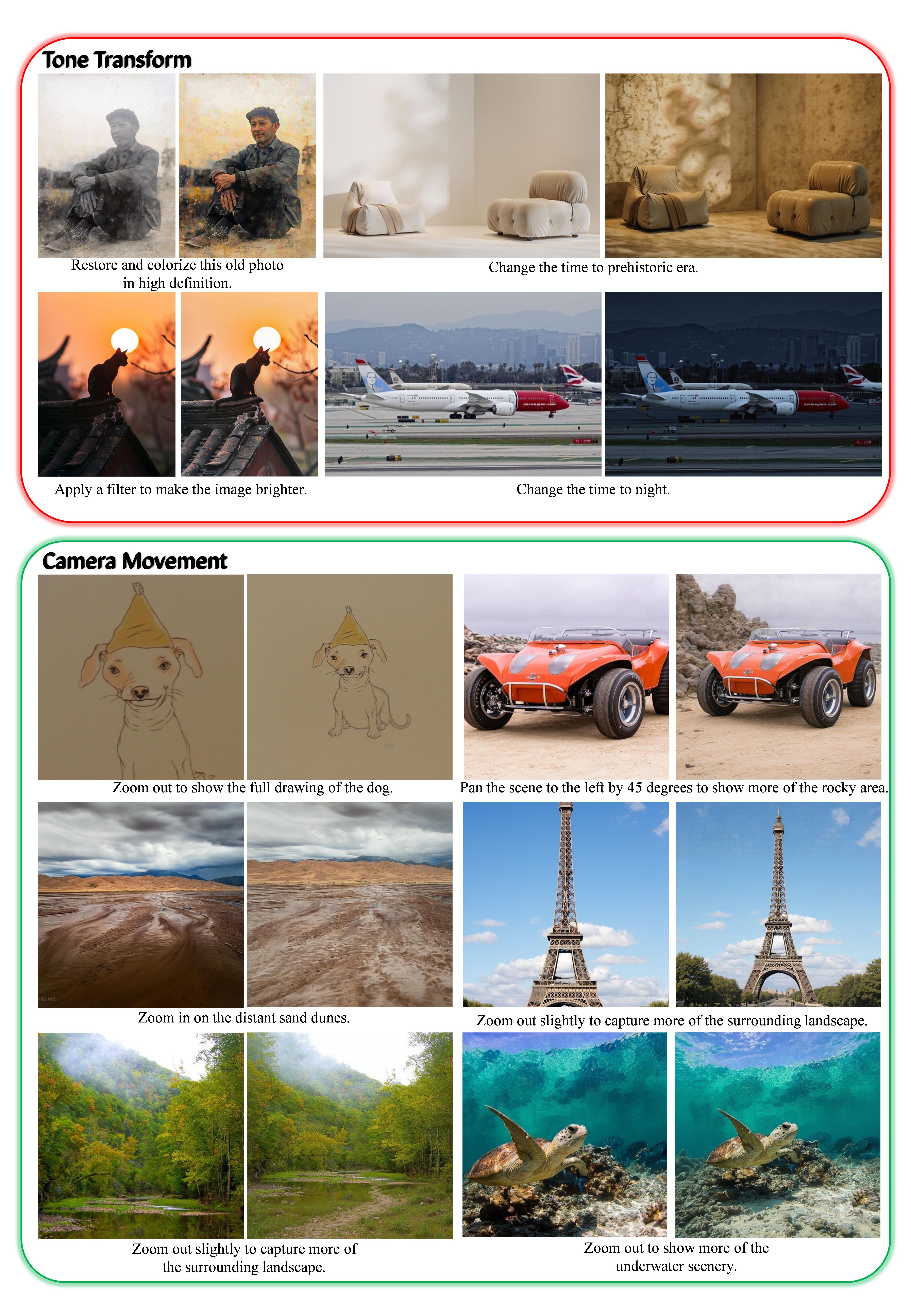}
\caption{Samples of X2Edit generation.}\label{generation 5}
\end{figure*}

\begin{figure*}[h]
\centering
\includegraphics[width=0.97\textwidth]{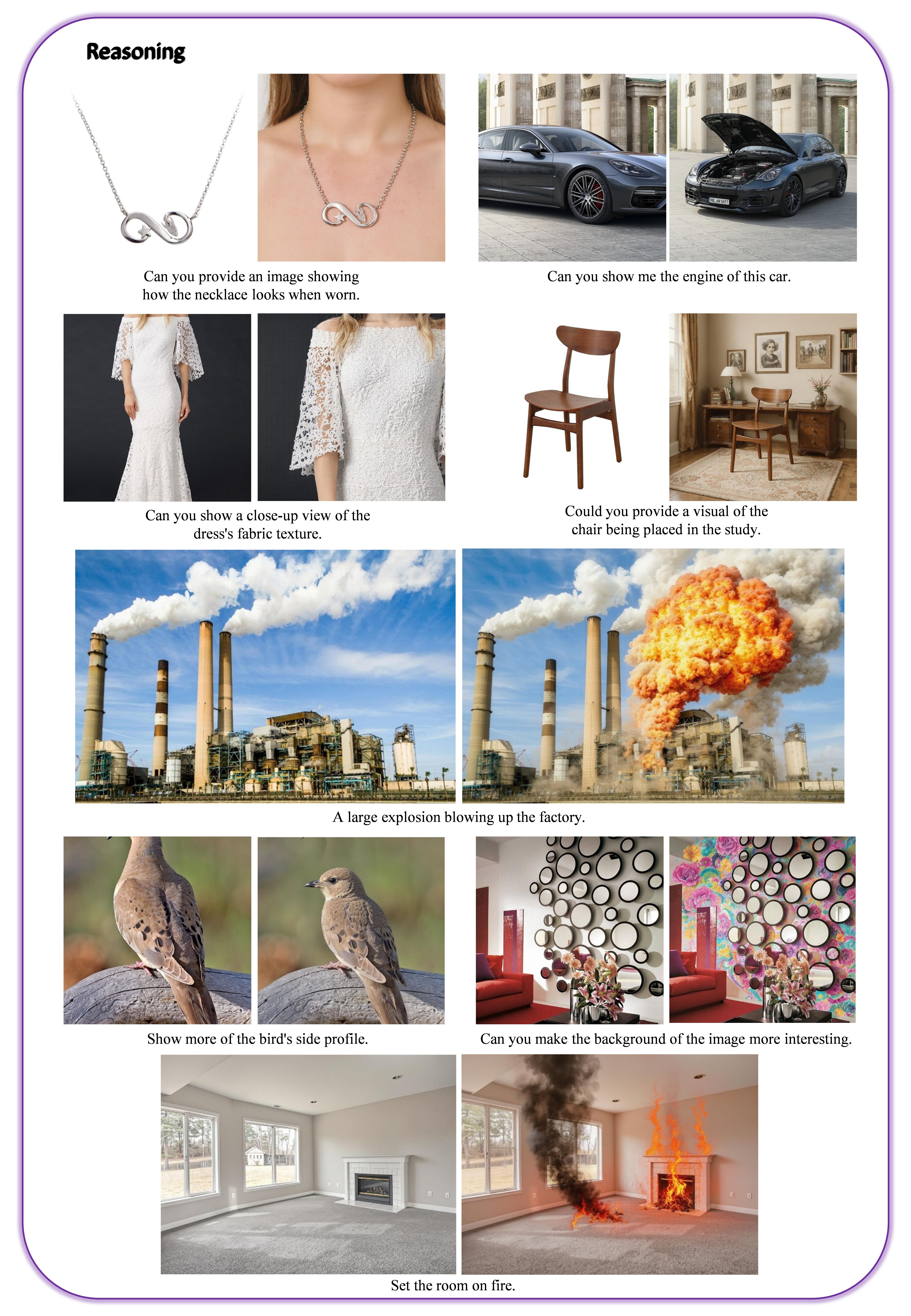}
\caption{Samples of X2Edit generation.}\label{generation 6}
\end{figure*}

\begin{figure*}[h]
\centering
\includegraphics[width=0.97\textwidth]{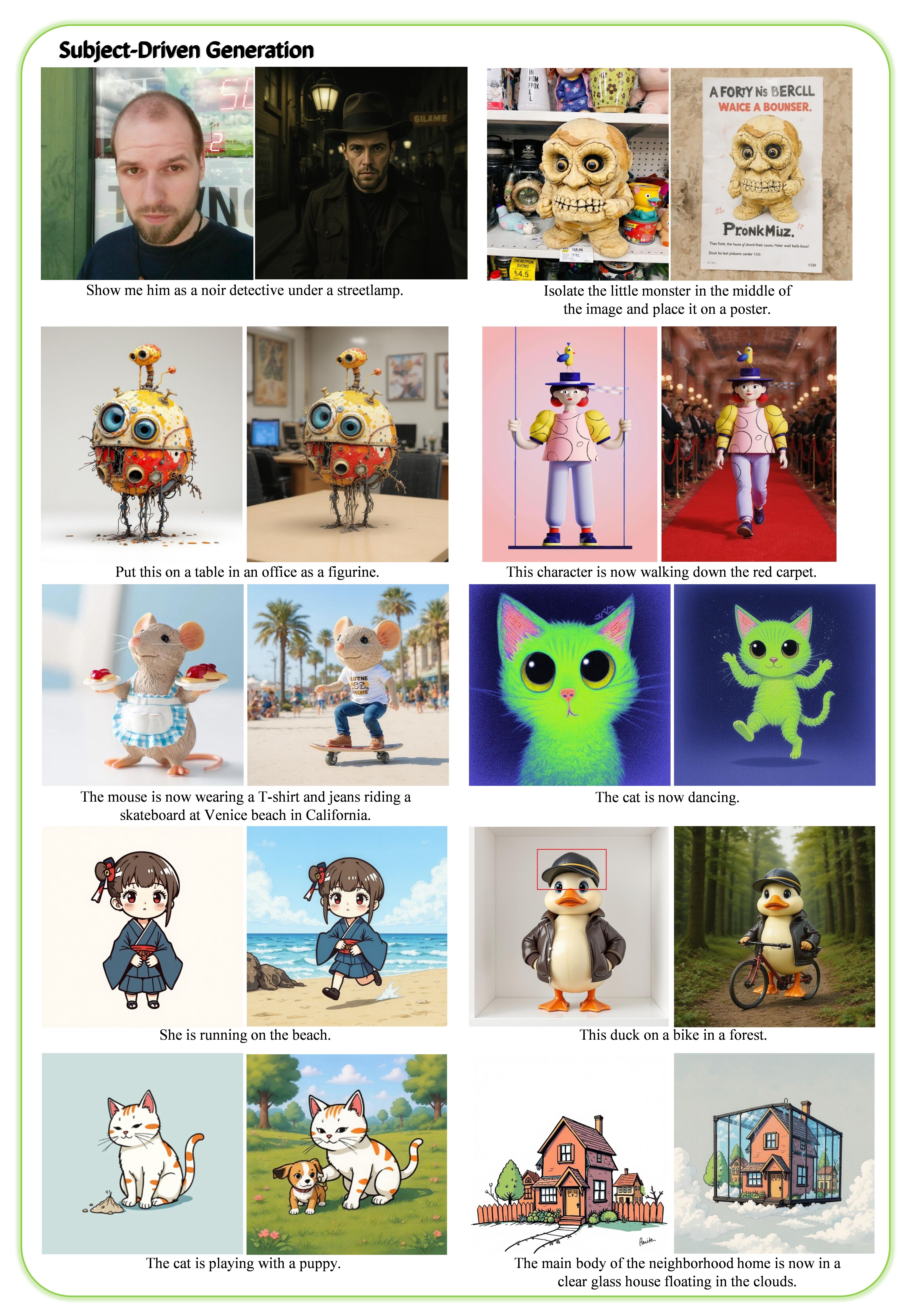}
\caption{Samples of X2Edit generation.}\label{generation 7}
\end{figure*}

\begin{figure*}[h]
    \centering
    \includegraphics[width=0.99\textwidth]{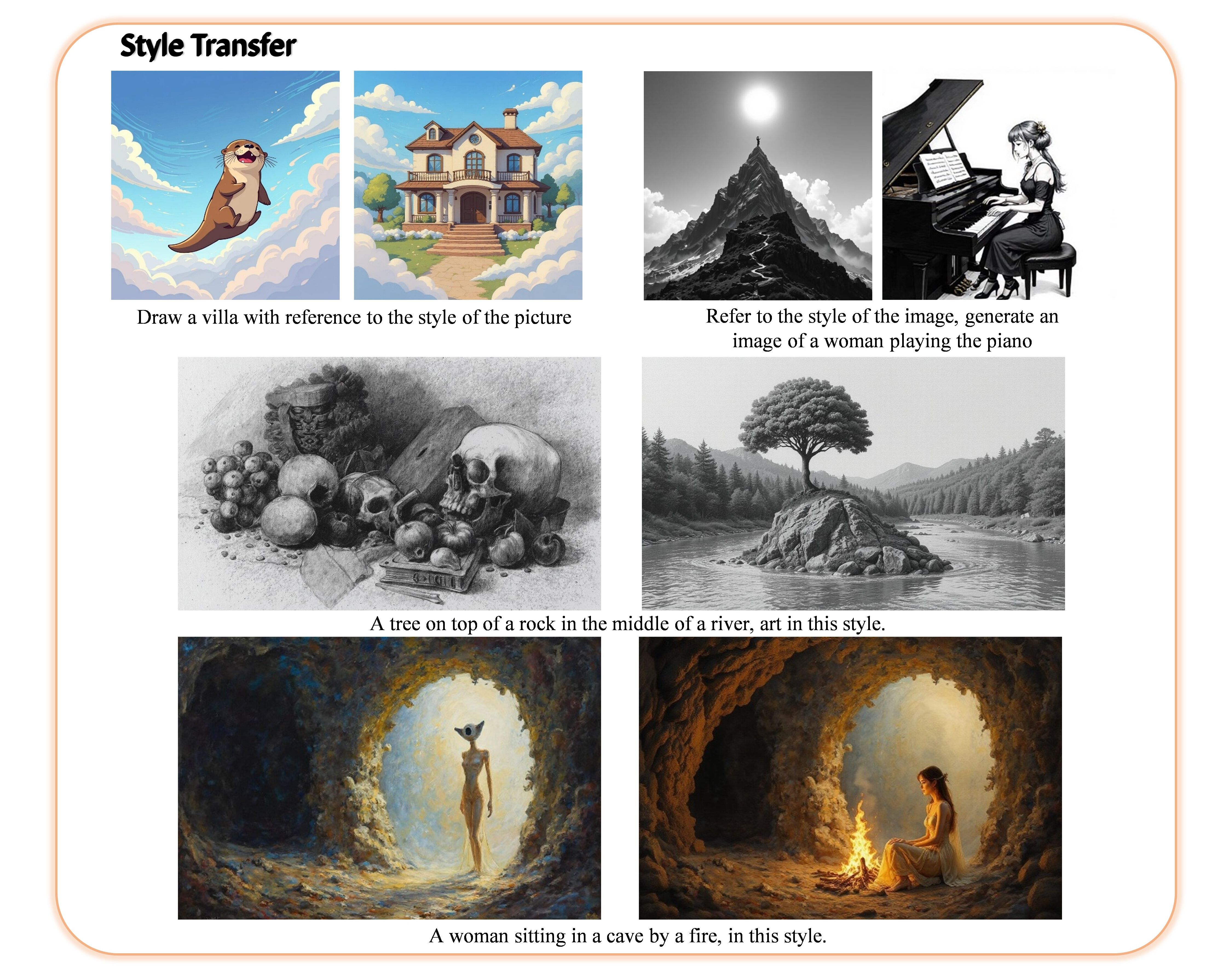}
    \caption{Samples of X2Edit generation.}\label{generation 8}
    
    \includegraphics[width=0.85\textwidth]{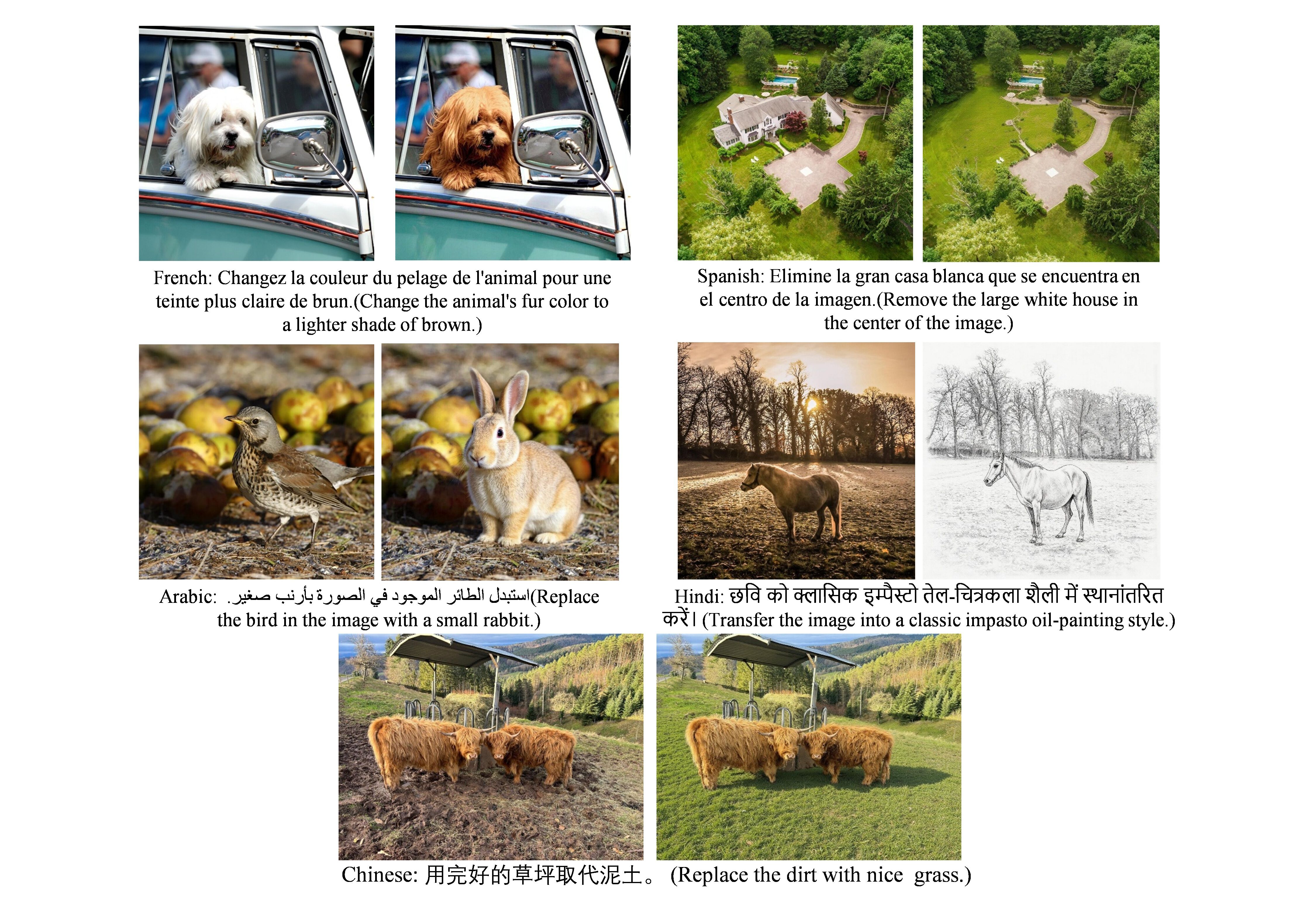}
    \caption{Samples of X2Edit generation under multiple language instructions.}\label{multi-languages}
\end{figure*}

Figure~\ref{generation 1}-\ref{generation 8} present editing results of X2Edit generation across all 14 editing tasks. Figure~\ref{multi-languages} demonstrates the editing results of X2Edit across five linguistically diverse editing instructions (Chinese, Hindi, Arabic, French, Spanish). These results demonstrate the powerful editing capabilities of X2Edit across multiple tasks and languages.

\section{Visual Comparison}
\label{Visual Comparison}

\begin{figure*}[h]
\centering
\includegraphics[width=0.92\textwidth]{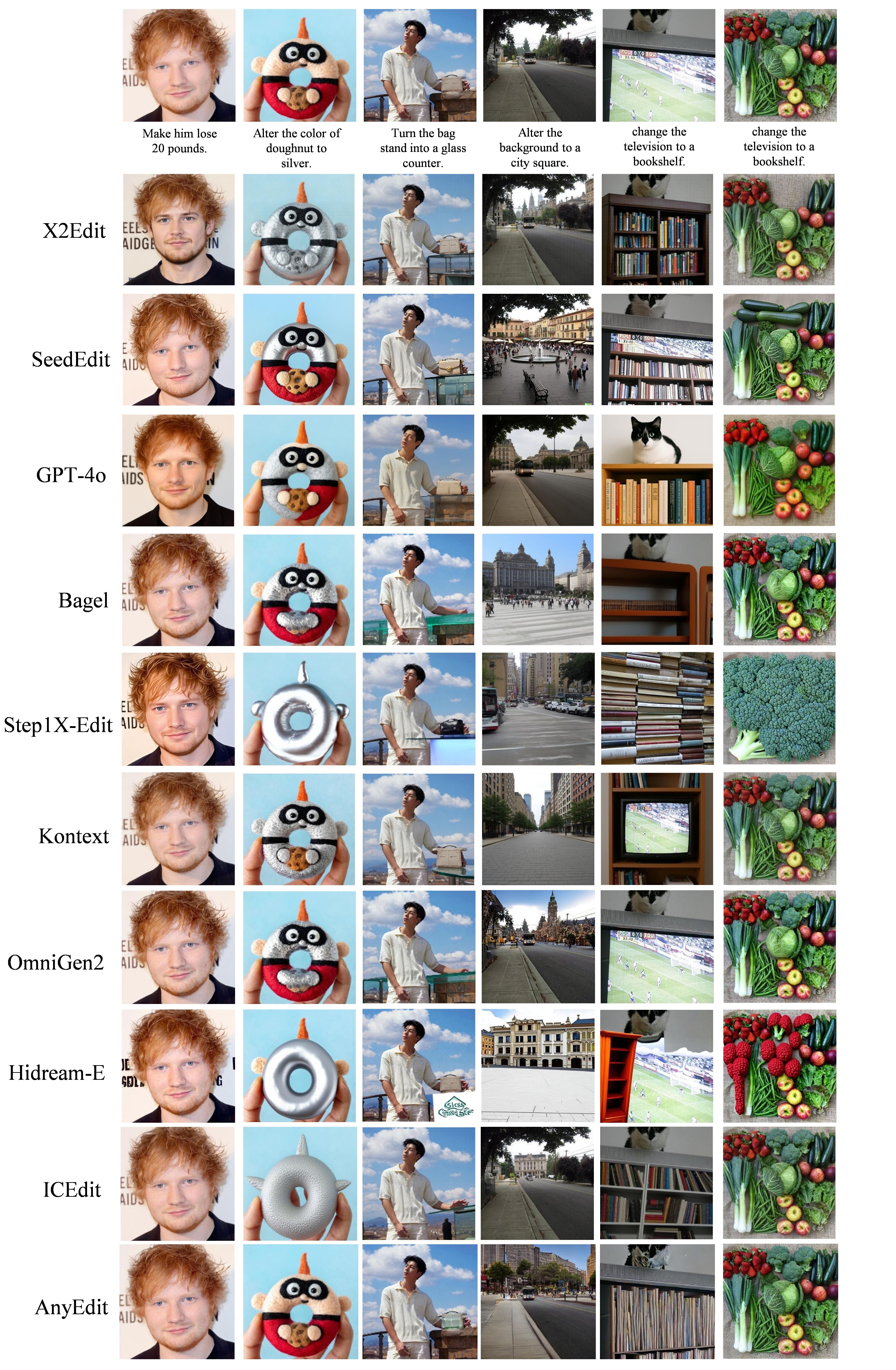}
\caption{Visual comparison of X2Edit and other comparative methods}\label{comparison 1}
\end{figure*}

Figure~\ref{comparison 1} presents comparative results of X2Edit and other main-stream methods.

\section{Limitations}
\label{Limitations}

\begin{figure*}[h]
\centering
\includegraphics[width=1\textwidth]{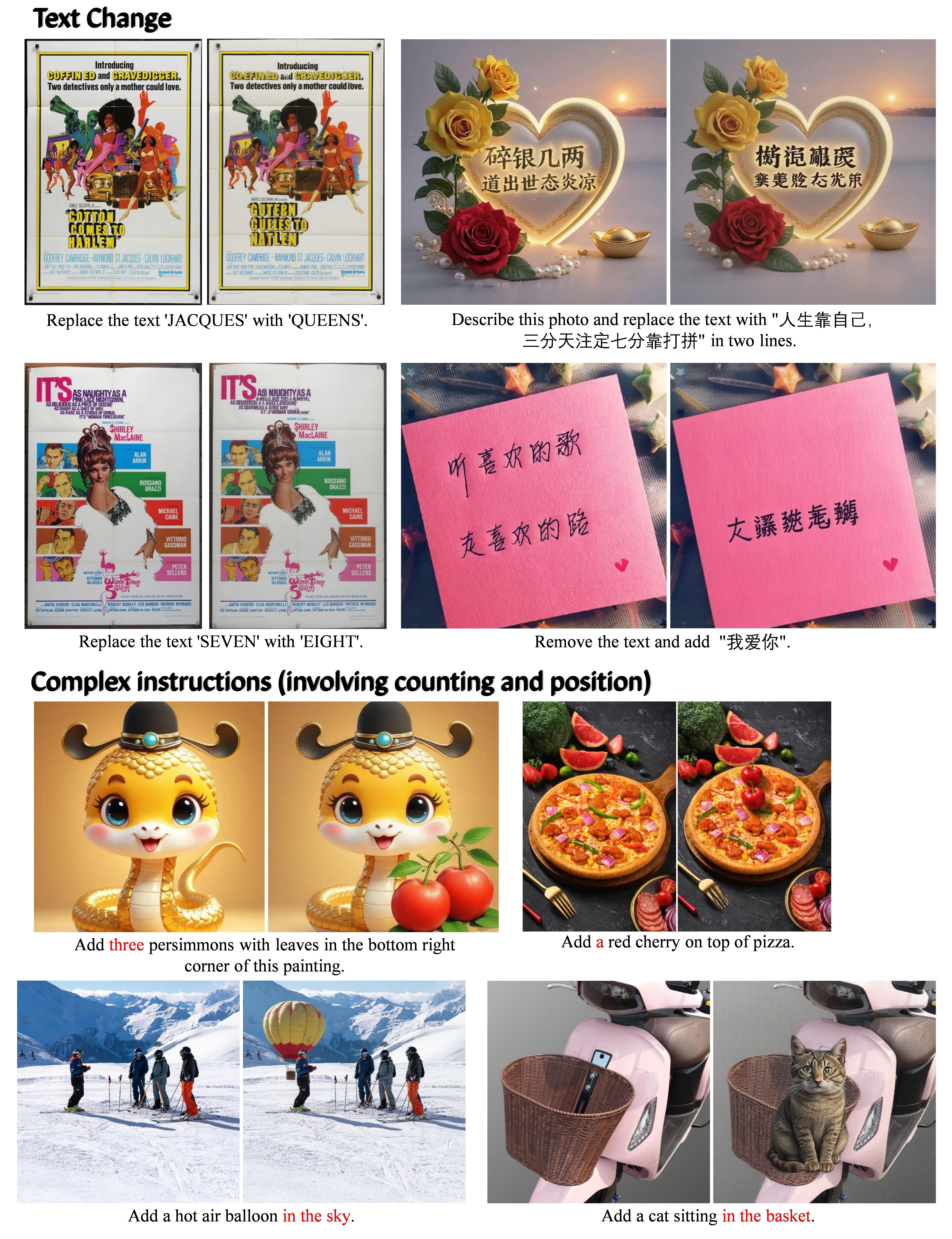}
\caption{Editing failure samples of X2Edit}\label{limitations}
\end{figure*}

X2Edit exhibits certain limitations in text change and execution of complex instructions. Figure~\ref{limitations} demonstrates failure cases in these two aspects. For text change, X2Edit struggles to accurately localize target text for editing. Moreover, constrained by the foundation model's inherent lack of Chinese text generation capability, X2Edit's Chinese text change capability remains limited despite LoRA fine-tuning. Additionally, X2Edit struggles to comprehend semantically complex instructions involving quantitative or spatial descriptions, resulting in superficial edits confined to basic visual semantics. Future research should focus on finer-grained image editing with deeper visual concept understanding and enhancing the model's text change capability.

\end{document}